\documentclass{article}
\usepackage{arxiv}

\usepackage[utf8]{inputenc} 
\usepackage[T1]{fontenc}    
\usepackage{hyperref}       
\usepackage{url}            
\usepackage{booktabs}       
\usepackage{amsfonts}       
\usepackage{nicefrac}       
\usepackage{microtype}      
\usepackage{xcolor}         
\usepackage{bbm}

\usepackage{graphicx,wrapfig,lipsum}
\usepackage{caption}
\usepackage{subcaption}
\usepackage{amssymb}
\usepackage{mathtools}

\def\eg{\textit{e.g.,~}}

\def\ie{\textit{i.e.~}}

\def\etal{\textit{et~al.~}}

\title{Learning from Long-Tailed Data with Noisy Labels}

\author{
{\bf Shyamgopal Karthik} \\
IIIT Hyderabad \\
India \\
\and
{\bf J\'erome Revaud}\\
Naver Labs Europe\\
France \\
\and
{\bf Chidlovskii Boris}\\
Naver Labs Europe\\
France
}

\begin{document}

%


\maketitle

\begin{abstract}
Class imbalance and noisy labels are 
the norm rather than the exception 
in many large-scale classification datasets. 
Nevertheless, most works in machine learning typically assume balanced and clean data. 
There have been some recent attempts to tackle, on one side, the problem of learning from noisy labels and, on the other side, learning from long-tailed data. 
Each group of methods make simplifying assumptions about the other.
Due to this separation, the proposed solutions often underperform
when both assumptions are violated. 
In this work, we present a simple two-stage approach based on recent advances in self-supervised learning to treat both challenges simultaneously . 
It consists of, first, task-agnostic self-supervised pre-training, followed by task-specific fine-tuning using an appropriate loss.
Most significantly, we find that self-supervised learning approaches are effectively able to cope with severe class imbalance. 
In addition, the resulting learned representations are also remarkably robust to label noise, when fine-tuned with an imbalance- and noise-resistant loss function.
We validate our claims with experiments on CIFAR-10 and CIFAR-100 augmented with synthetic imbalance and noise, as well as the large-scale inherently noisy Clothing-1M  dataset. 
\end{abstract}

\section{Introduction}
\label{sec:introduction}
Starting with the seminal work of Krizhevsky~\etal~\cite{krizhevsky2012imagenet}, Deep Neural Networks (DNNs) have been remarkably successful when trained under supervision of large-scale labeled data. 
However, this success has hinged upon two strong yet implicit assumptions: (i) data is balanced, \ie there are equal number of samples for all categories; and (ii) 
all annotated labels are clean and reliable. 
In practice, unfortunately, these assumptions are incredibly difficult and expensive to meet. 
In fact, the price to collect and annotate by human annotators a large-scale dataset such as ImageNet is immense~\cite{liao21annotating}. 
Conversely, it is now clear that collecting large-scale datasets can be cheap and fast when affording to violate these two assumptions~\cite{babenko14neural, karpathy14large, li2017webvision, mahajan18exploring, sun17revisiting}.
It is therefore desirable to conceive learning algorithms that can handle imbalance and noise \emph{simultaneously}.

In this paper, we address the problem of learning from imbalanced data, also referred to as long-tailed class distributions, in the presence of label noise. 
It is well-known that class imbalance and noisy labels both pose significant challenges.
Consequentially, a vast amount of research has looked into mitigating the impact of these aspects separately. Recent methods coping with noisy labels apply different techniques ranging from sample selection~\cite{han18coteaching,jiang18mentornet} to label correction~\cite{arazo2019unsupervised,patrini2017making} as well as noise-aware losses~\cite{castells2020superloss,han2021survey,liu2020early,wang2021learning}. 
To learn effectively from long-tailed distributions, some works have proposed to modify the sampling algorithm to ensure all classes are represented equally~\cite{chawla2002smote,kubat1997addressing}, modify the loss function~\cite{menon2020logitadj}, or perform a post-hoc correction~\cite{kang2020decoupling}. 
However, existing methods designed to learn from noisy labels assume a balanced class distribution and, conversely, methods tailored to learn from long-tailed class distributions assume labels to be clean. 

We argue that such a separation is artificial since label noise and long-tailed class distributions occur simultaneously in real-world datasets. 
For instance, the Clothing1M dataset~\cite{xiao15clothing1m}, collected automatically from shopping websites, contains an estimated amount of 38.5\% incorrect labels and its most populated class contains almost 5 times more instances than the smallest one (this ratio is denoted as the \emph{class imbalance ratio}). 
Other examples include the Landmarks dataset~\cite{babenko14neural}, a large-scale image retrieval dataset, and the WebVision dataset~\cite{li2017webvision}, an inexpensive analogue of ImageNet, that were both collected in a semi-automatic manner using image search engines:
they contain respectively around 75\% and 20\% annotation errors~\cite{gordo2016deep,li2017webvision} and have imbalance ratios of about $10^4$ and 24, respectively.
\begin{wrapfigure}{r}{5.5cm}
\includegraphics[width=5.5cm]{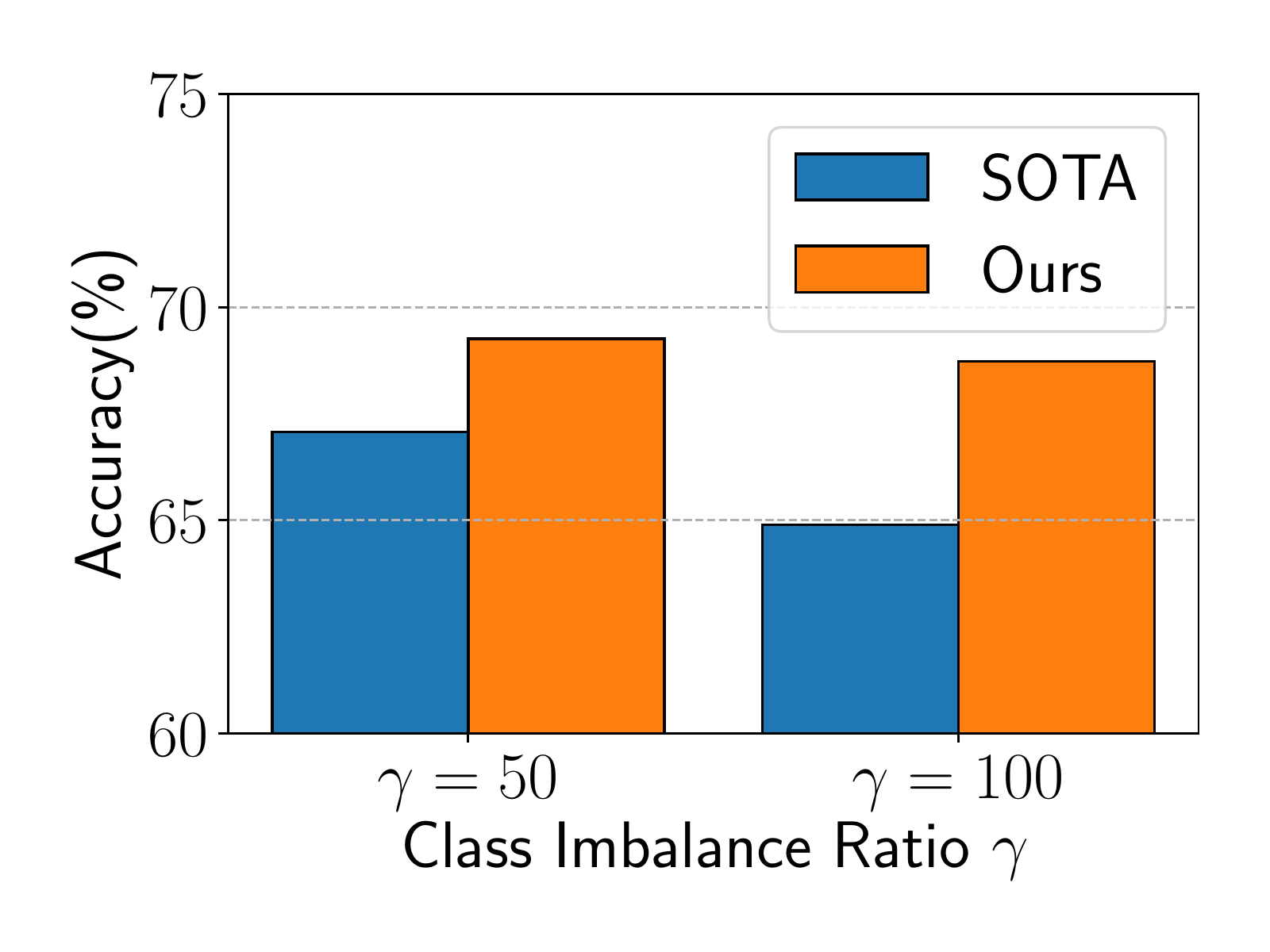}
\caption{Our method significantly outperforms the state-of-the-art (SOTA) on Clothing-1M when the training set is severely imbalanced. The class imbalance ratio $\gamma$ denotes the ratio of the largest class size over the smallest one.}
\label{wrap-fig:1}
\end{wrapfigure} 


In this paper, we show that the existing methods tailored to learn from noisy labels degrade drastically in the presence of long-tailed class distributions.  
To address both class imbalance and label noise in deep learning, we propose to split the training procedure in two stages: representation learning and classifier training. We first pre-train the model in a self-supervised manner by discarding the training labels. This is followed by fine-tuning the model (\ie learning a robust classifier) using the noisy and long-tailed class labels.  

This procedure is inspired by recent findings in semi-supervised learning where it was discovered that self-supervised pre-training leads to state-of-the-art performance \cite{chen2020simclr2}.
Likewise, we pre-train the model to be invariant to image augmentations in the first stage. We experiment with diverse formulations of recently proposed self-supervised learning methods, namely SimCLR~\cite{chen2020simclr}, Barlow Twins~\cite{zbontar2021barlowtwins}, BYOL~\cite{grill2020byol} and SimSiam~\cite{chen20simsiam}. 
We find that all these methods are able to learn high-quality representations even when the samples are drawn from strongly imbalanced distributions. 

In the second stage, the self-supervised model is fine-tuned using the noisy labels. To cope with class imbalance and noise, we adopt a simple solution based on two strong baselines.
Namely, we show that a combination of the Logit Adjustment loss~\cite{menon2020logitadj}, a classification loss adapted to long-tailed data, and the SuperLoss~\cite{castells2020superloss}, a generic loss robust to label noise, can be used to fine-tune a classifier that is robust to both class imbalance and label noise. 
Our main contributions can be summarized as follows:
\begin{enumerate} 
    \item Our results suggest that, perhaps surprisingly, self-supervised approaches are robust to imbalanced data. In particular, we experiment with a diverse set of self-supervised techniques and observe very similar outcomes in all cases.
    \item We propose a simple and effective strategy to cope with both class imbalance and noisy labels. It is composed of two stages: (1) self-supervised pre-training and (2) supervised finetuning with two complementary losses that respectively prevents the model from being biased towards the dominant classes and reduce the memorization of noisy labels.
    \item Under both label noise and long-tailed setting, our approach outperforms state-of-the-art methods by large margins on CIFAR as well as the Clothing-1M dataset (see Figure~\ref{wrap-fig:1}).

\end{enumerate}

\section{Method}
\label{sec:method}

Our approach is inspired by recent advances in semi-supervised learning~\cite{chen2020simclr2}. It leverages available data in, first, task-agnostic and, second, task-specific ways. 
Given an image dataset with possible class imbalance and noisy labels, we first use an augmentation invariance criterion to pre-train a model in a self-supervised manner. 
Second, we fine-tune this representation using a loss function tailored for long-tailed data and noisy labels. Below we first review existing techniques for self-supervised training and then describe the two stages of our approach in detail. 

\subsection{Background on Augmentation-invariant Self-Supervision}
\label{ssec:sota_ssl}

Prior work has shown that self-supervision for visual data can be tackled in various ways.
In contrast to older approaches that propose a variety of pretext tasks~\cite{zhang2016colorful,gidaris2018unsupervised,pathak2016context}, recent approaches all revolves around the principle of learning invariance to random image augmentations (\eg scaling, color jitter, blur, etc.) using a Siamese network architecture~\cite{caron2020swav,chen2020simclr,chen20simsiam,grill2020byol,he2020moco,zbontar2021barlowtwins}. 
Specifically, the goal is to maximize the similarity between the encoded representations of two augmented versions of the same image.
Because this procedure can collapse to trivial solutions, different remedies have been proposed. 

Earlier methods like SimCLR~\cite{chen2020simclr}, SimCLRv2~\cite{chen2020simclr2} and MoCo~\cite{he2020moco}, for instance, use negative samples and contrastive losses based on artificially constructed positive and negative pairs. More recent methods like BYOL~\cite{grill2020byol} and SimSiam~\cite{chen20simsiam} have shown that it is possible to use only positive pairs. 
The trick is to rely on asymmetric operations such as propagating gradients through only one of the two Siamese branches.
SimSiam~\cite{chen20simsiam}, for its part, can be seen as a minimalist version of BYOL~\cite{grill2020byol} without momentum encoder.
Clustering-based method like SwAV~\cite{caron2020swav}, DeepCluster~\cite{caron2018deep} or Sela~\cite{asano2019self} are instead based on trainable versions of the $k$-means algorithm that learns image representations leading to clusters stable against random augmentations.
Lastly, Barlow Twins~\cite{zbontar2021barlowtwins} is able to prevent collapse without considering image pairs at all nor asymmetric operations thanks to a novel loss function based on redundancy reduction. 
While these approaches appear very diverse, they all display excellent performance for visual representation learning in situations where data is perfectly balanced (typically, on ImageNet~\cite{russakovsky15imagenet}).
In this work, we  show that the benefits of self-supervised pre-training extend to the imbalanced setting as well.

\begin{wrapfigure}{r}{4.7cm}
\vspace{-2mm}
\includegraphics[width=4.7cm]{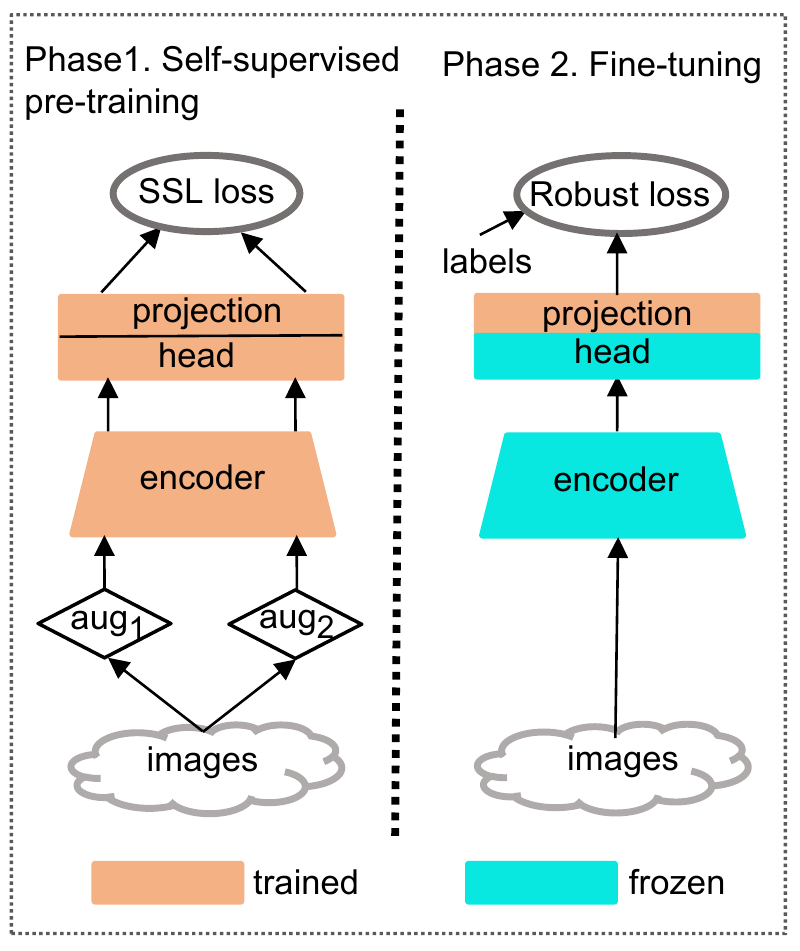}
\caption{Summary of the proposed two-staged approach: (1) self-supervised model pre-training; and (2) fine-tuning the projection head with a robust loss.
}
\label{wrap:2stages}
\end{wrapfigure} 

\subsection{Self-Supervised Pre-training Stage} 
\label{ssec:ssl}
The first stage of our approach consists of pre-training the model in a self-supervised manner, thereby discarding the instance labels.
Since one of our goals is to study the impact of class-imbalanced distributions on different self-supervised techniques, we select a diverse subset of those methods.
Namely, we experiment with SimCLR
(uses positive and negative pairs), BYOL
(only positive pairs and momentum encoder), SimSiam 
(same as BYOL but no momentum encoder) and Barlow 
(none of the previous). 
Perhaps surprisingly, as recent attempts to explain the success of these self-supervised approaches assume batch-level balanced data~\cite{purushwalkam020demystifying}, we find that all of them work well even under severe class imbalance (see Section~\ref{sec:evaluation}).

\textbf{Training pipeline.}
All these methods use Siamese networks where each image $x$ is augmented twice. The two augmented views are fed to an {\it encoder} network (a ResNet~\cite{he16resnet}) and then transformed with a MLP {\it projection head} composed of 2 or 3 fully-connected (FC) layers. 
During training, obtained representations are fed to the contrastive loss in the case of SimCLR or to a redundancy reduction loss for Barlow Twins, and to a prediction head followed by similarity losses in the cases of BYOL and SimSiam (see Supplementary material for more details.) 
\subsection{Fine-tuning Stage}
\label{ssec:finetune}

Fine-tuning is a common way to adapt a task-agnostic pre-trained network for a specific task, which in our case consists of learning with noisy labels. 
We follow recently proposed strategies for semi-supervised learning~\cite{chen2020simclr2,chen20simsiam,grill2020byol}.
In more detail, we freeze the entire encoder network during finetuning and only train the MLP projection head. 
The head can be trained entirely, or only partially.
In the latter case, we fine-tune the model from a middle layer of the projection head.

\textbf{Loss functions.}
We leverage two recently proposed loss functions during finetuning to ensure robustness against both label noise and class imbalance. 
These are chosen for their simplicity and effectiveness, as well as for the fact that they can be easily combined.
The first one, the logit adjustment loss~\cite{menon2020logitadj}, is a modified version of the Cross-Entropy loss that can handle class imbalance.
Given a model $f_\theta$ and  observed class distribution $\pi_y$ which predicts logits $f(x)$ for a sample $x$, the logit adjustment corrects the logits as follows:
\begin{equation}
     f^{*}(x) = f(x) + \log(\pi_y).  
\end{equation}
Taking softmax over the adjusted logits, the cross-entropy loss can be applied for classification of $x$:
\begin{equation}
     L_{\text{LA}} = - \log\left(\frac{\exp({f^*_y(x)})}{\sum_{y'} \exp(f^*_{y'}(x))}\right). 
\label{eq:ce}
\end{equation}

The second loss is the SuperLoss~\cite{castells2020superloss}, a generic loss for curriculum learning.
Even though this loss is not primarily meant to handle noise, recent works suggest that curriculum learning has strong noise-resistant abilities~\cite{castells2020superloss,wu2020curricula}. 
Its effect is to downweight the contribution of hard samples (\ie those having a higher loss value), effectively preventing the memorization of noisy labels.
Given the loss $L_{\text{LA}}$ from Eq.~\eqref{eq:ce}, the SuperLoss computes a new loss as follows:
 \begin{equation}
     L_{\text{LA+SL}}(L_{\text{LA}},\sigma^*) = (L_{\text{LA}} - \tau)\sigma^* + \lambda(\log\sigma^*)^2,
 \end{equation}
where $\lambda$ is a regularization trade-off and $\sigma^*$ corresponds to a per-sample confidence whose optimal value can be computed in closed-form as:
 \begin{equation}
     \sigma_{\lambda}^{*}(L_{\text{LA}}) = \exp\left[-W\left(\frac{1}{2}\max\left(\frac{L_{\text{LA}} - \tau }{\lambda},\frac{2}{e}\right)\right)\right], \label{eq:sigma}
\end{equation}
where $W$ stands for the Lambert $W$ function, $\tau$ is the expected loss for the ``average'' sample and is used to separate the easy samples from the hard samples.

\section{Experiments/Results}
\label{sec:evaluation}
\subsection{Datasets}
\label{ssec:datasets}

We evaluate the proposed methodology on two standard benchmarks with simulated label noise and class imbalance, CIFAR-10 and CIFAR-100, and one large-scale, real-world dataset, Clothing1M. 
CIFAR datasets consist of $32\times32$ color images composed of 10 and 100 classes, respectively. Each dataset contains 50,000 train and 10,000 test images. For both CIFAR datasets, we simulate label noise by replacing the labels for a certain fraction of the training samples with labels chosen from a uniform distribution. 
We also conduct experiments on Clothing1M, which contains 14 classes
with 
1 million $256\times256$ train images collected from online shopping websites with labels generated using surrounding text. The proportion of annotation errors is estimated at 38.5\%~\cite{xiao15clothing1m}. 

\noindent \textbf{Class imbalance.} 
Following prior work on class imbalance, we create imbalanced versions of CIFAR-10 and CIFAR-100 and down-sample the number of samples per class by following the Exponential profile~\cite{cao2019learning,cui19class} with imbalance ratio $\gamma = max_y p(y)/min_y p(y)$. 
We test our method on CIFAR-10 with $\gamma=50$ and $\gamma=100$. For $\gamma=100$, the smallest class then comprises $5000/\gamma=50$ training instances. 
When we further apply 90\% of label noise, only 5 of them keep their original labels while the other 45 get their labels switched to random ones (possibly keeping the same labels). 
Note that the proportion of incorrect labels for CIFAR-10 needs to be strictly less than 90\%, as under 90\% actual noise the true labels cannot possibly be recovered~\cite{li2020dividemix}.
In the case of CIFAR-100, we test our method with $\gamma=5$ and $\gamma=10$. In the latter case, the smallest class has $500/\gamma=50$ instances, which matches the case of CIFAR-10 with $\gamma=100$. 
For Clothing-1M, which has 14 classes, we also experiment with long-tailed versions with $\gamma=50$ and $\gamma=100$.


\subsection{Implementation Details}
\label{ssec:implementation}
We briefly describe the self-supervised training protocols for each considered method. In all cases, we stay as close as possible to the original protocols, only making minimal changes required to pass from ImageNet-based training to CIFAR and Clothing1M.

\noindent \textbf{Backbone encoder.} 
For all the datasets, we use a ResNet18~\cite{he16resnet} backbone. For the CIFAR-10 and CIFAR-100, the first convolutional layer has a stride of $3\times3$ instead of the usual $7\times7$ and the first max pooling layer is removed~\cite{he16resnet}. 
During SSL pre-training and fine-tuning, we attach  a projection head whose configuration depends on the specific SSL method, as described below. 

\textbf{SimCLR:} We pre-train SimCLR for 1000 epochs using the Adam optimizer~\cite{kingma2014adam} with a learning rate of~$0.001$, a weight decay of $10^{-6}$ and a batch size of $512$. For fine-tuning the model on the noisy labels, we train a linear classifier on top of the representations extracted by the encoder. For this, we train for 25 epochs using Adam with a learning rate of~$0.001$ and a weight decay of $10^{-6}$.

\textbf{SimSiam:} We follow the original SimSiam implementation, except that we use 2 FC layers in the projection head instead of 3 as we find it slightly better in our case.
We pre-train the network for 800 epochs using SGD with a learning rate of $lr \times bs/256$, a base $lr=0.03$ and a batch size $bs=512$. We use a cosine decay schedule except during the initial warm-up period where it is scaled linearly for 10 epochs. 
The weight decay is set to 0.0005 and the SGD momentum is 0.9. 
For fine-tuning, we train the full projection head (2 FC layers) with Adam for 10 epochs and a learning rate of 0.003 without weight decay and with a batch size of 256. When the noise exceeds $\nu_{\text{SimSiam}}=60\%$, we only fine-tune the last FC layer with a learning rate of $0.01$. This strategy is very similar the method proposed in SimCLRv2~\cite{chen2020simclr2}. 

\textbf{BYOL:} We follow the original architecture for the projection and predictor heads~\cite{grill2020byol}, but we use the Adam optimizer instead of LARS~\cite{you2017large} due to the small size of the CIFAR training sets. 
We use the same pre-training protocol as for SimSiam, with a base learning rate set to 0.001 and a weight decay of $1.5\cdot 10^{-6}$.
For fine-tuning, we again follow SimSiam, but this time switching from 2 FC layers to 1 FC layer at $\nu_{\text{BYOL}}=40\%$ noise.

\textbf{Barlow Twins:} As for BYOL, we pre-train using the Adam optimizer instead of LARS as in~\cite{zbontar2021barlowtwins}. The pre-training protocol is identical to SimSiam except the base learning rate is set to 0.003. The $\lambda$ parameter is kept to $0.005$ as in~\cite{zbontar2021barlowtwins} but we find that setting the size of the projection head's hidden and output layers to 2048 improves performance.
For fine-tuning, we again use the same protocol as for BYOL and SimSiam except for $\nu_{\text{BarlowTwins}}=20\%$ noise.

\subsection{Baselines} 
\label{ssec:baselines}

We compare our approach to several baselines and state-of-the-art approaches.
First, we compare against strong baselines consisting of training a network from scratch in a single stage using one of the aforementioned specialized losses: the Logit-Adjusted loss~\cite{menon2020logitadj}, the SuperLoss~\cite{castells2020superloss}, and our proposed combination of both losses in order to deal jointly with class imbalance and label noise.
For reference, we also compare against a standard Cross-Entropy baseline.

Finally, we compare against DivideMix~\cite{li2020dividemix} and ELR~\cite{liu2020early}, two state-of-the-art methods which have both shown excellent robustness to label noise. 
Note that these methods require significant modifications compared to the standard learning procedure used by baselines and our approach, such as network ensembling, weight averaging and mix-up data augmentation~\cite{zhang2017mixup}. 
We use their default implementations available, and we adapt these to the long-tailed settings. 

\subsection{CIFAR experiments}
\label{ssec:results}

\begin{figure*}[t]
\centering
\begin{subfigure}{0.245\textwidth}
\centering
\includegraphics[width=\linewidth]{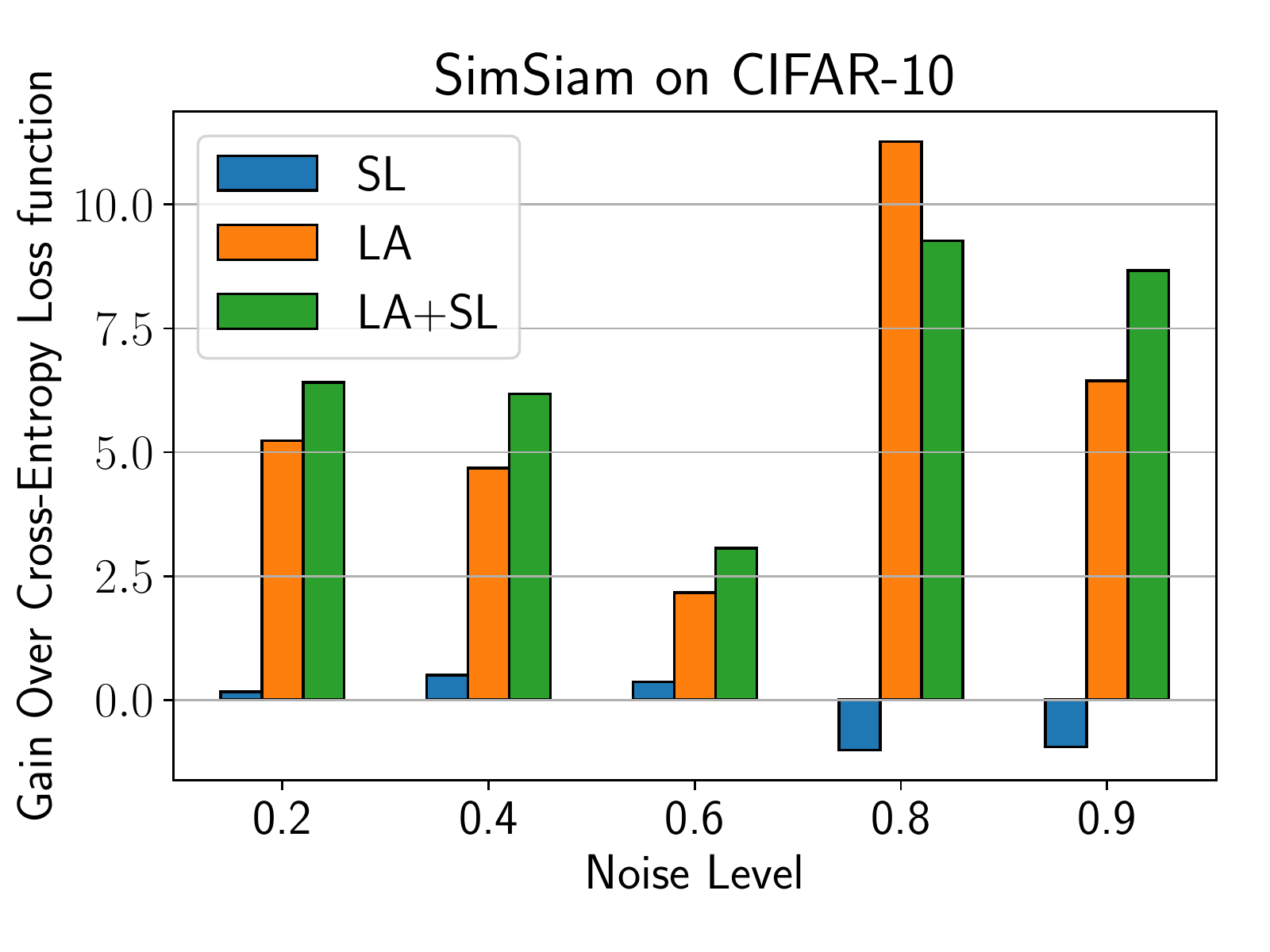}
\end{subfigure}
\begin{subfigure}{0.245\textwidth}
\centering
  \includegraphics[width=\linewidth]{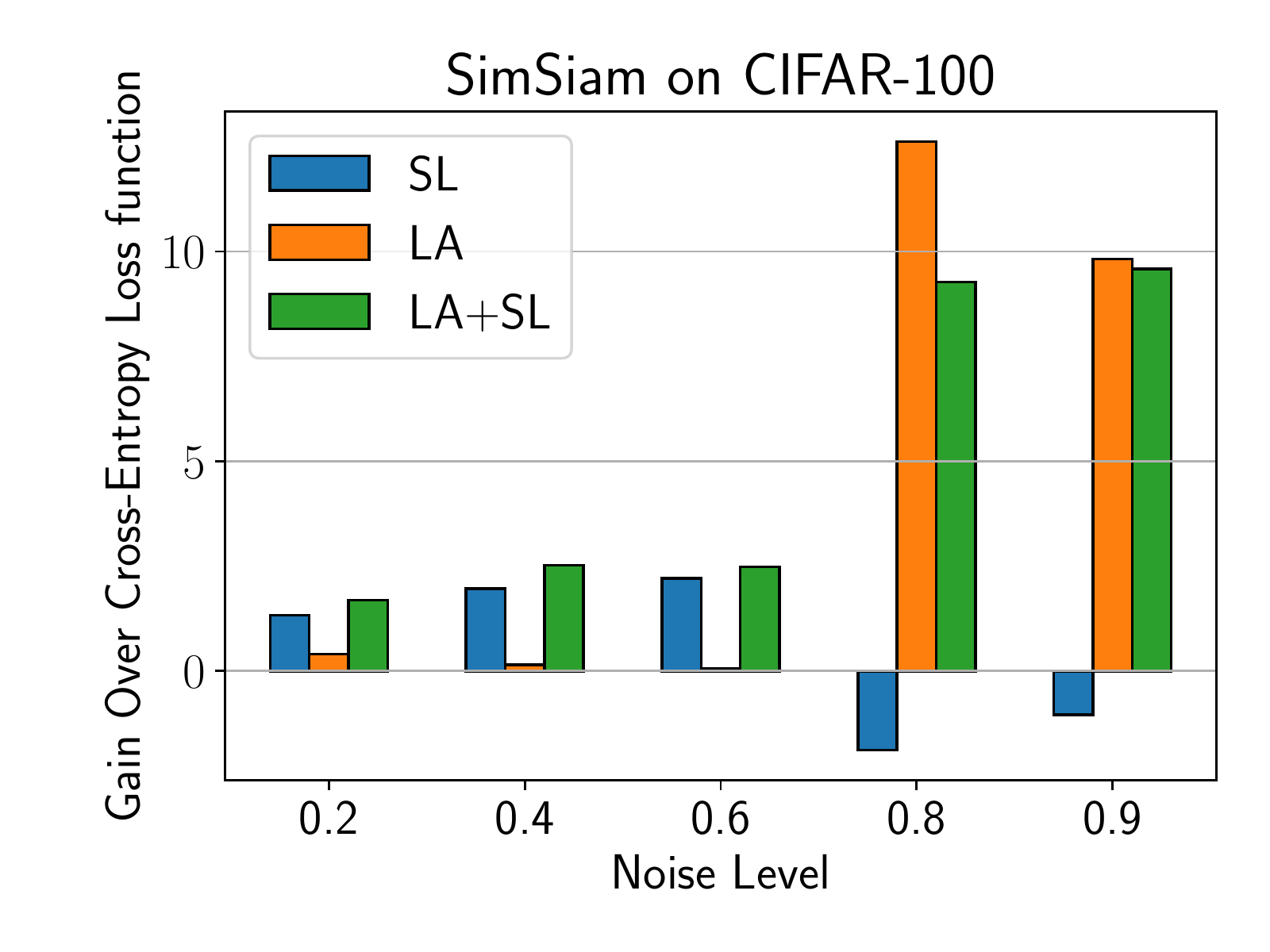}
\end{subfigure}
\centering
\begin{subfigure}{0.245\textwidth}
\centering
  \includegraphics[width=\linewidth]{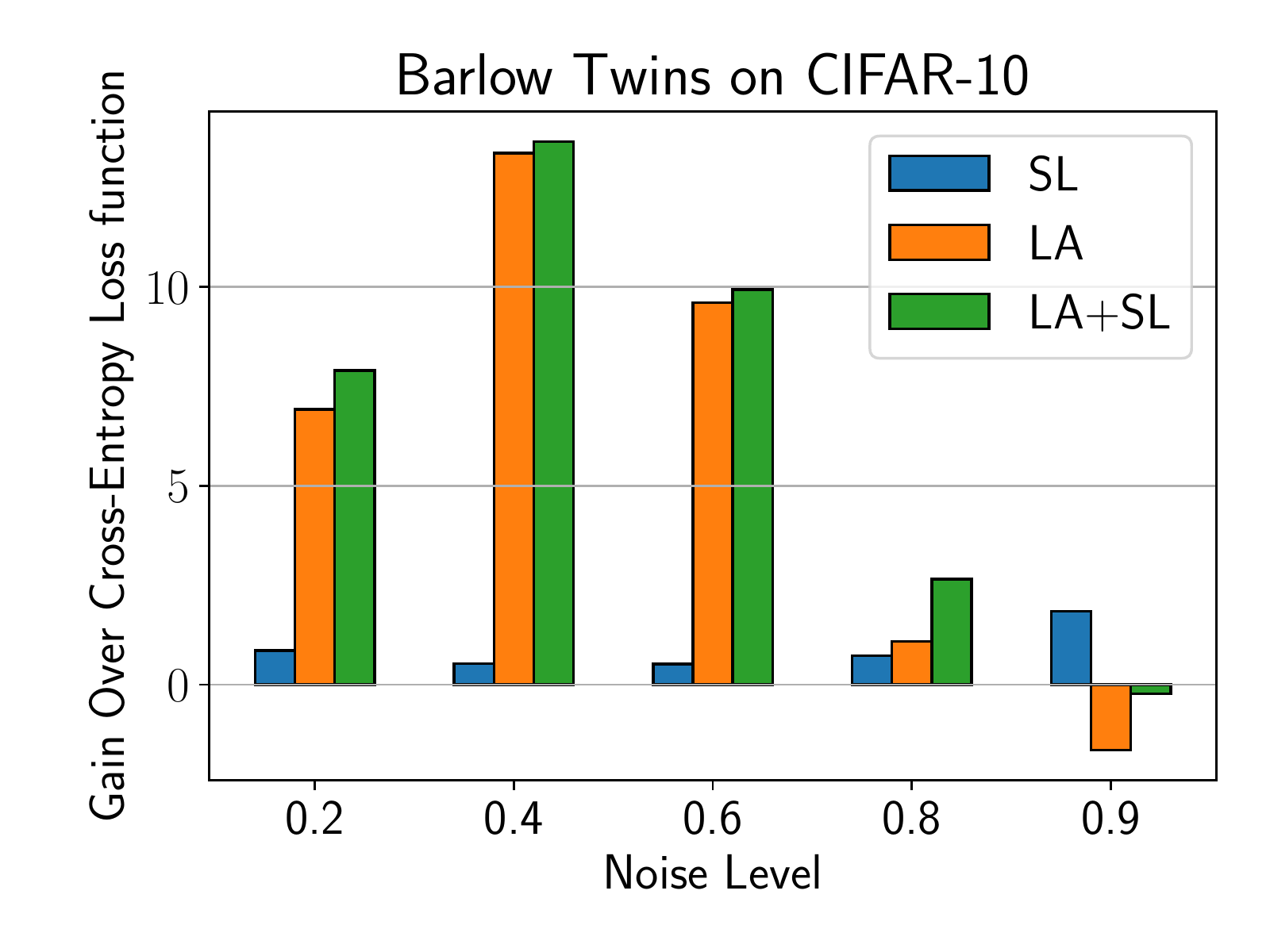}
\end{subfigure}
\begin{subfigure}{0.245\textwidth}
\centering
  \includegraphics[width=\linewidth]{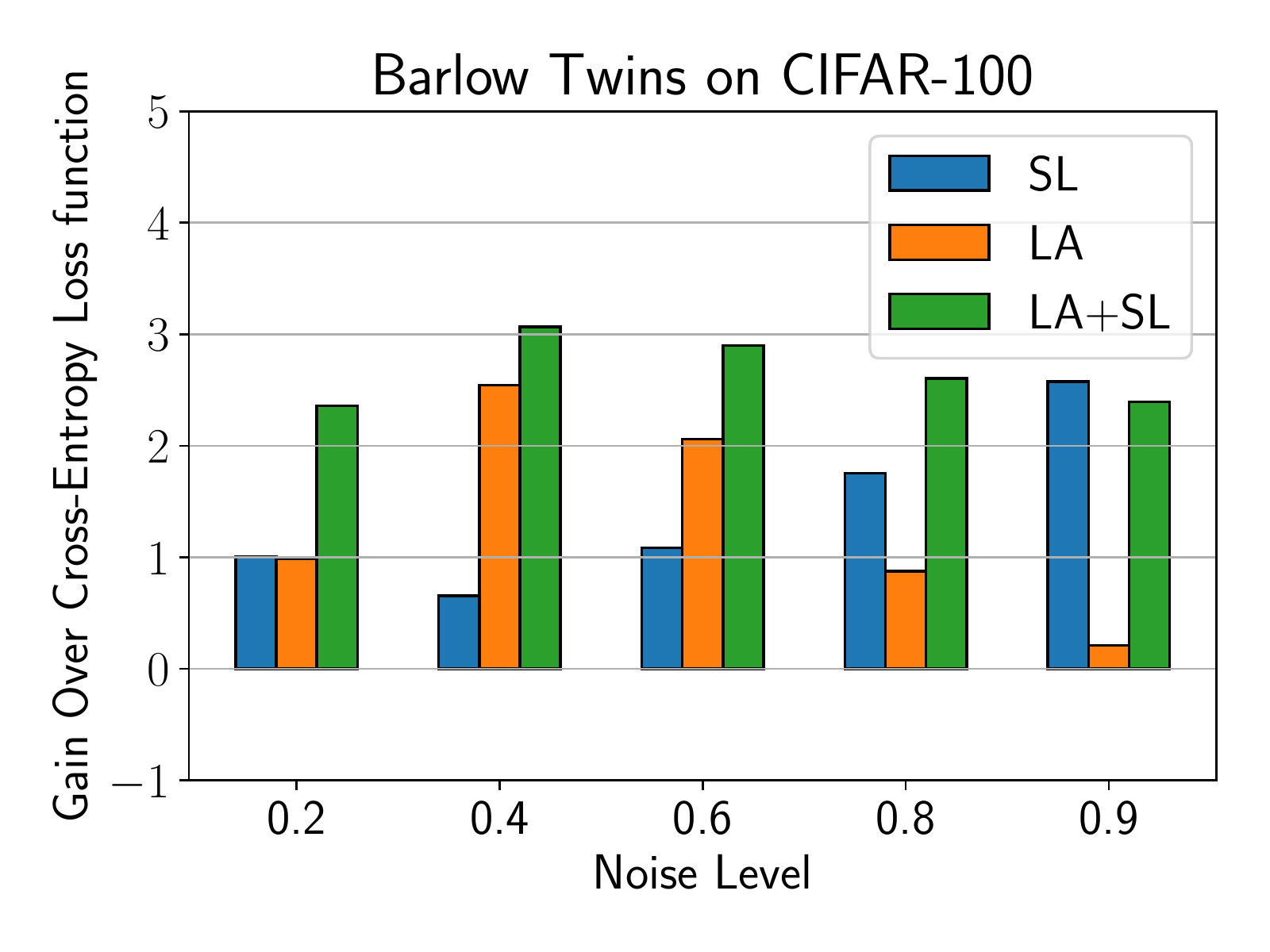}
\end{subfigure}

\caption{Accuracy gain compared to the cross-entropy fine-tuning baseline for models pretrained using SimSiam and Barlow Twins. Results are averaged over all imbalanced ratios (\ie $\gamma={50,100}$ for CIFAR-10 and $\gamma={5,10}$ for CIFAR-100).}
\label{fig:loss_plots}
\end{figure*}

\textbf{Fine-tuning losses.}
We first study the impact of the imbalance- and noise-tailored losses considered in Section~\ref{ssec:finetune} during finetuning of the two-stage learning process.
Namely, we consider the 4 following configurations: CE, CE+SL, LA, LA+SL where CE and LA respectively refers to the Cross-Entropy and Logit-Adjusted losses, and ``+SL'' denotes applying SuperLoss on top of another loss~\cite{castells2020superloss}.
For the SuperLoss hyper-parameters, we always use a fixed threshold $\tau=\log(C)$
and we set the regularisation parameter to $\lambda=4$. 
Results are presented in Figure~\ref{fig:loss_plots} in terms of absolute improvement compared to the CE loss for models pre-trained using SimSiam and Barlow Twins (BYOL and SimCLR pre-training yield similar outcomes). 
While we observe some variability depending on which self-supervised method is used, we find that combining LA+SL almost always achieves the best performance overall. 
The accuracy gain can reach over 10\%, which validates the effectiveness of these two losses.
Here, it is important to note that the gains from both LA and SL are essentially for free as it does not require any additional information nor additional training time.
In the remainder of this section, we therefore use LA+SL as default fine-tuning losses unless stated otherwise.

\begin{figure*}[t]
\centering
\begin{subfigure}{0.32\textwidth}
\centering
\includegraphics[width=\linewidth]{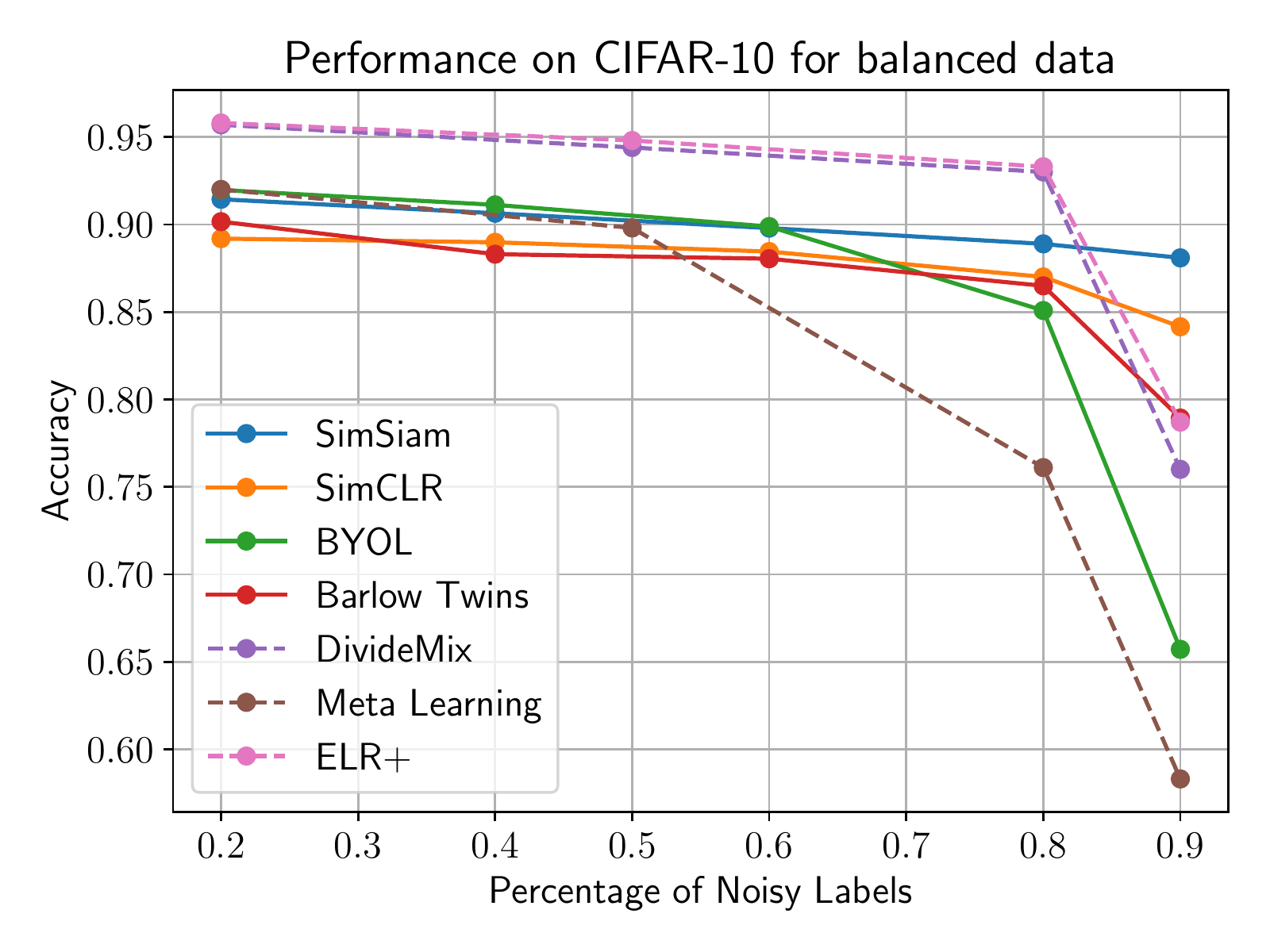}
\caption{}
\label{fig:top1_1}
\end{subfigure}
\begin{subfigure}{0.32\textwidth}
\centering
  \includegraphics[width=\linewidth]{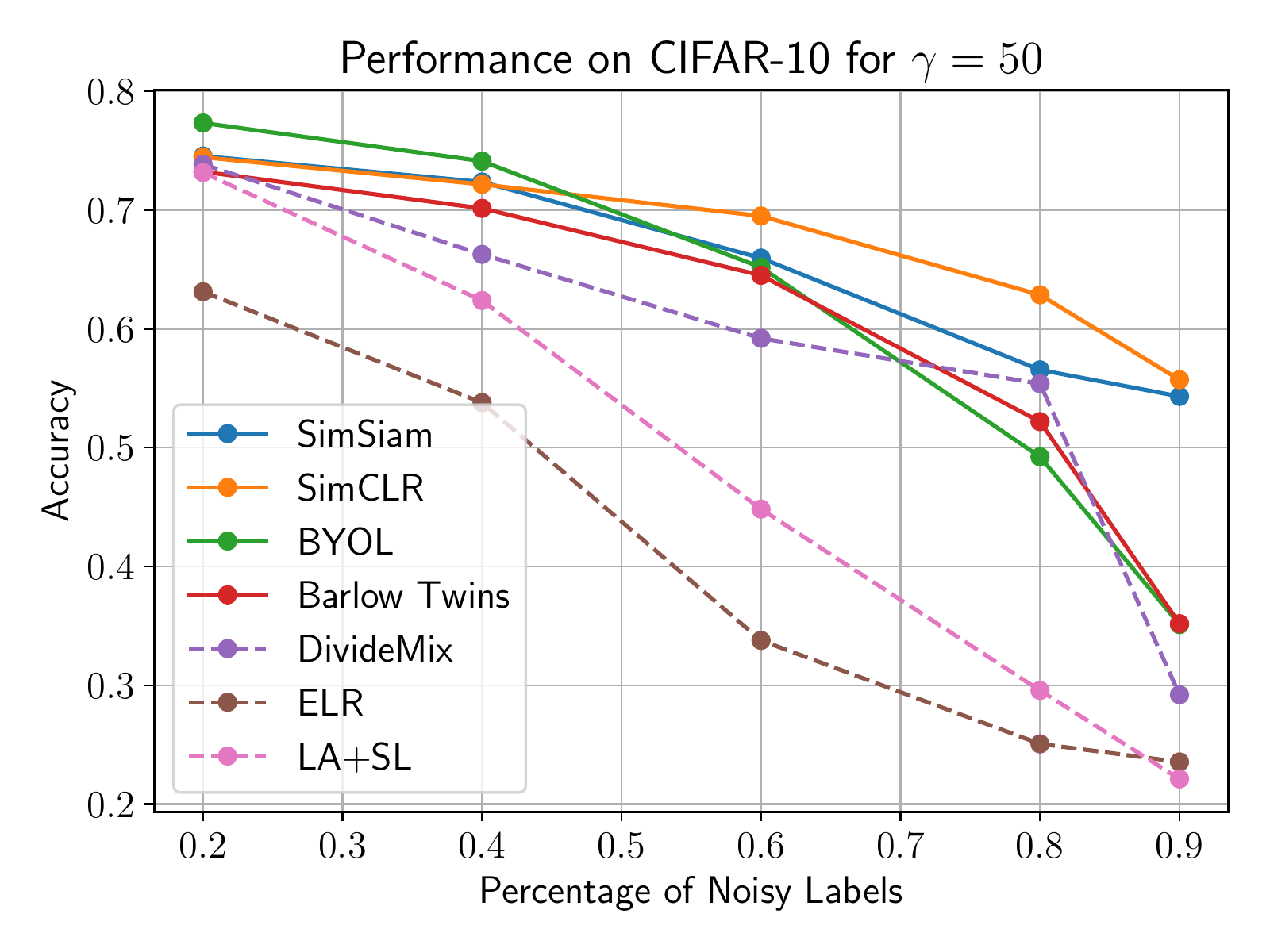}
  \caption{}
 \label{fig:top1_2}
\end{subfigure}
\begin{subfigure}{0.32\textwidth}
\centering
  \includegraphics[width=\linewidth]{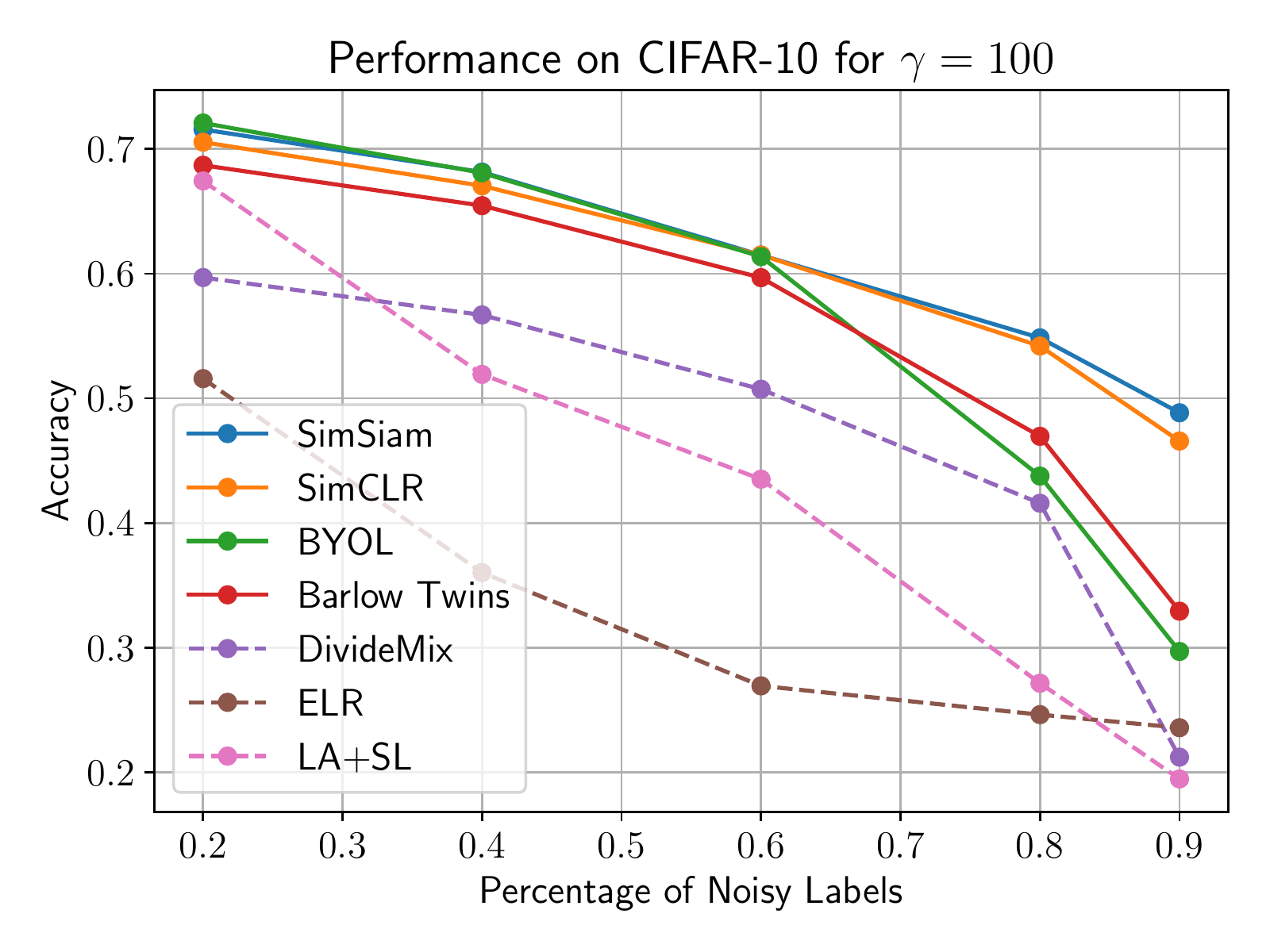}
  \caption{}
  \label{fig:top1_3}
\end{subfigure}
\newline
\begin{subfigure}{0.32\textwidth}
\centering
  \includegraphics[width=\linewidth]{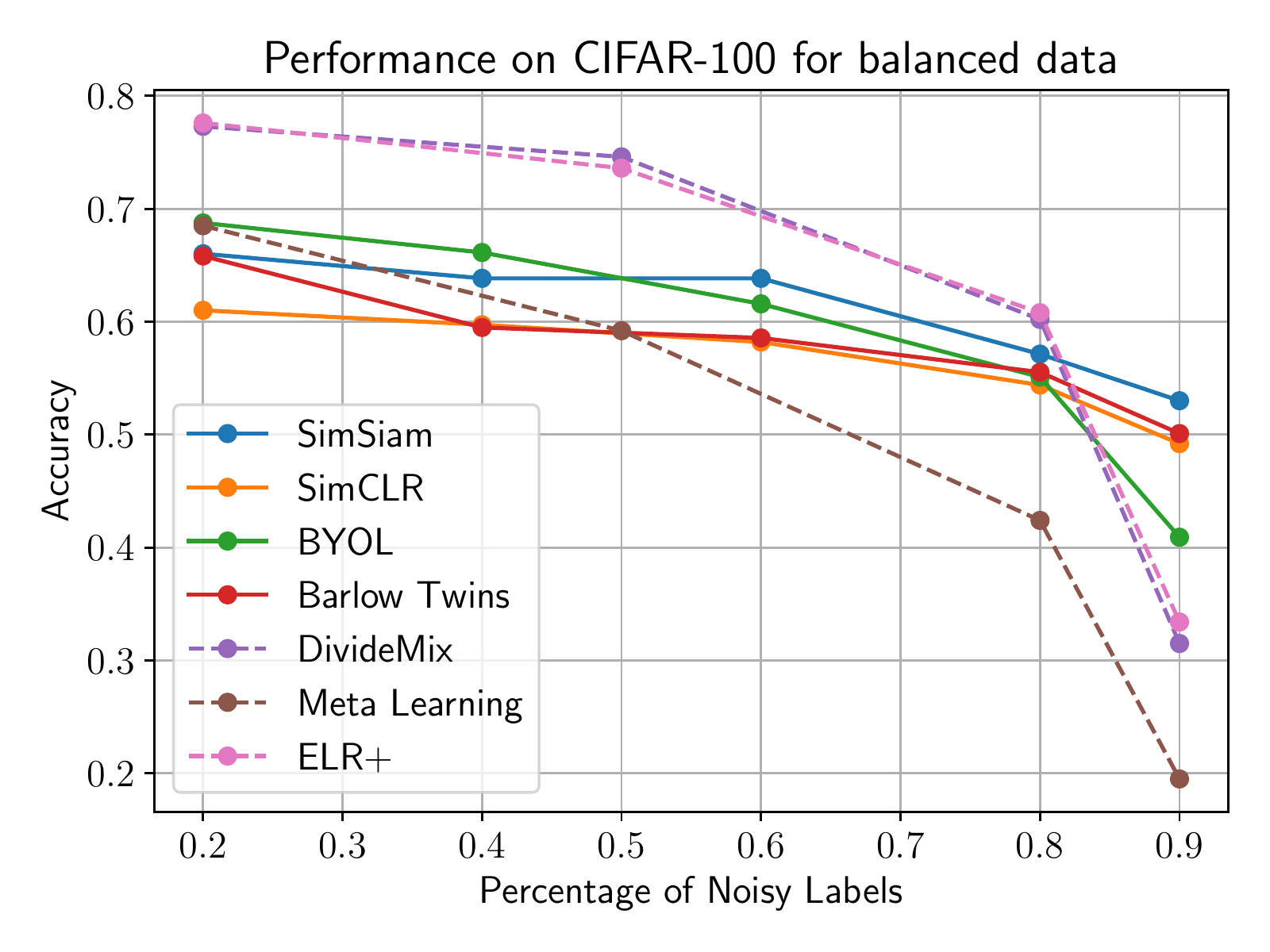}
\caption{}
\label{fig:top2_1}
\end{subfigure}
\begin{subfigure}{0.32\textwidth}
\centering
  \includegraphics[width=\linewidth]{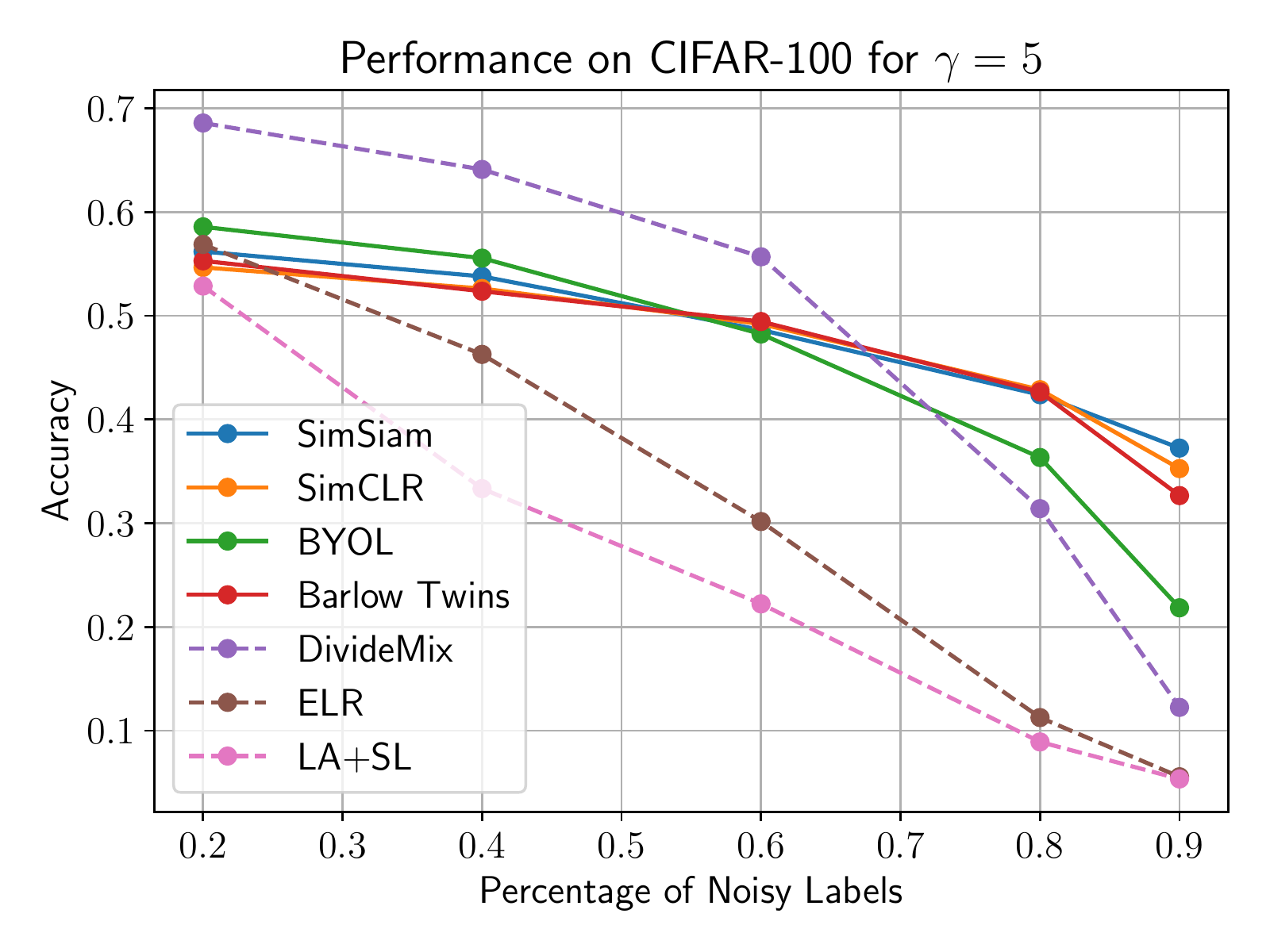}
\caption{}
\label{fig:top2_2}
\end{subfigure}
\begin{subfigure}{0.32\textwidth}
\centering
  \includegraphics[width=\linewidth]{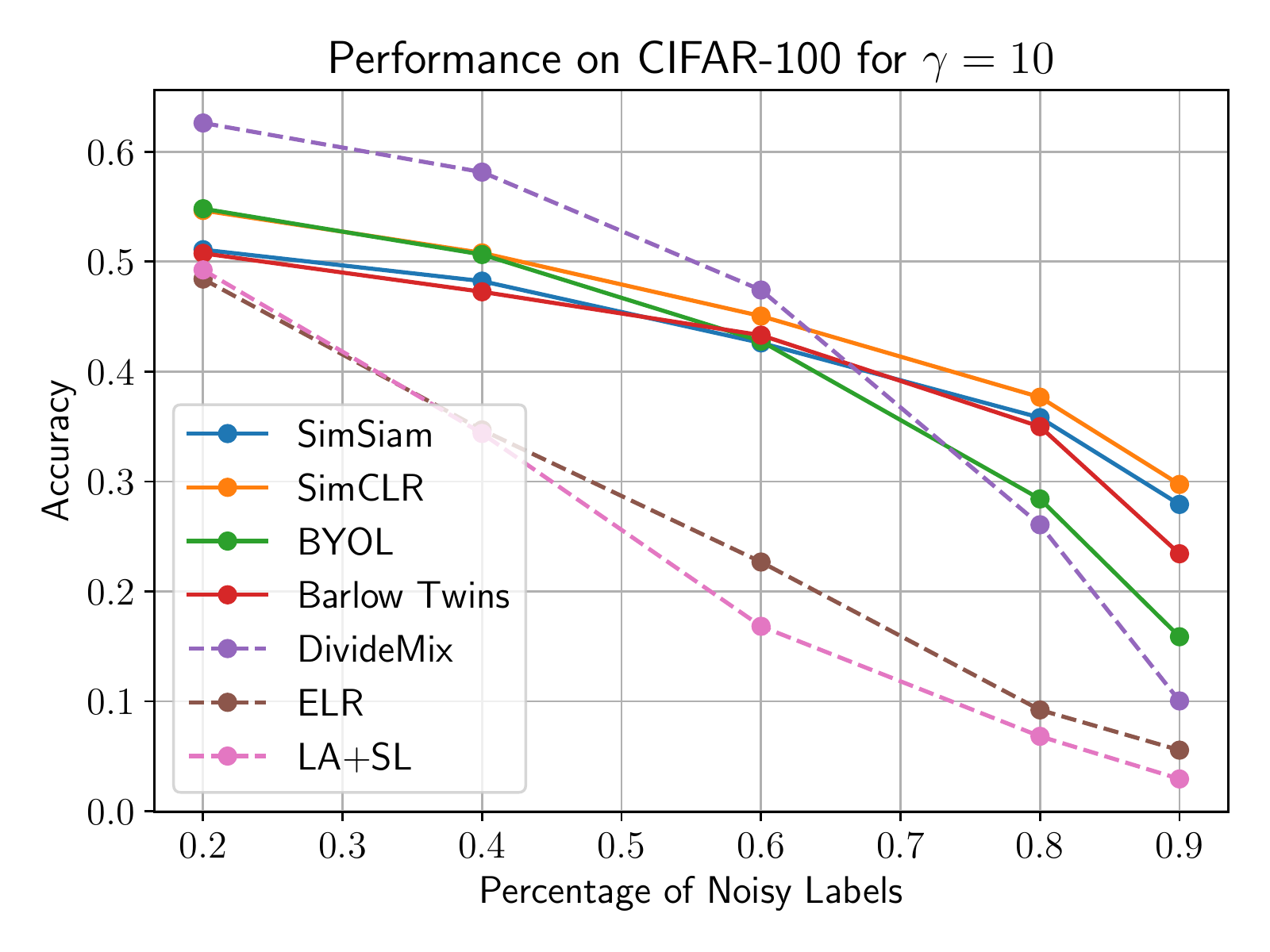}
  \caption{}
\label{fig:top2_3}
\end{subfigure}
\caption{
    Results on CIFAR-10 and CIFAR-100 datasets as a function of symmetric noise.
    }
\label{fig:cifar_plots}
\end{figure*}

\textbf{Comparison of self-supervised methods:}
We present experimental results for each of the considered self-supervised methods in Fig.~\ref{fig:top1_1}-\ref{fig:top1_3}
for CIFAR-10 and 
Fig.~\ref{fig:top2_1}-\ref{fig:top2_3}
for CIFAR-100.
We observe that BYOL significantly outperforms other methods in low-to-moderate noise levels by up to 3\textasciitilde4\%. This is consistent with the overall superior results achieved by BYOL in balanced and noiseless settings compared to other self-supervised methods~\cite{chen20simsiam,grill2020byol}. 
We hypothesize that this superiority is due to the weight-averaging trick used by BYOL, which is also employed to improve results by ELR~\cite{liu2020early}.
Other methods, in comparison, performs on par at this noise regime.
Conversely, we find that BYOL significantly underperforms compared to the other methods at high-noise levels. 
Interestingly, SimSiam, which is mostly similar to BYOL in  principle except for the weight-averaging part, is either on par or significantly better than other self-supervised approaches under severe noise.
Overall, we find it interesting and rather unexpected that all self-supervised approaches are robust to strong class imbalance. For instance, all approaches obtains at least 60\% accuracy on CIFAR-10 with $\gamma=100$ at 60\% noise, which is unprecedented.

\begin{figure*}[t]
\centering
\begin{subfigure}{0.245\textwidth}
\centering
\includegraphics[width=\linewidth]{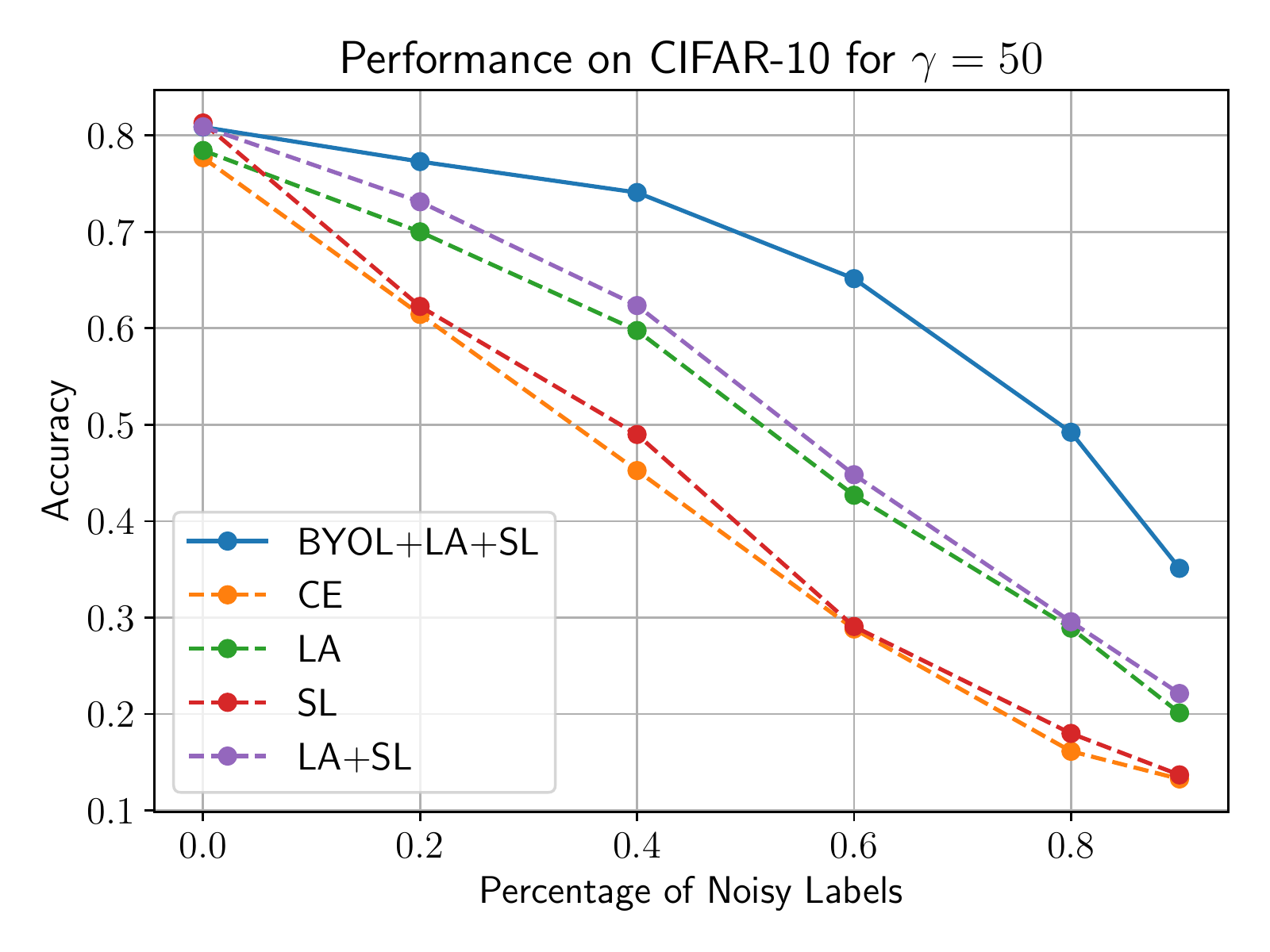}
\end{subfigure}
\begin{subfigure}{0.245\textwidth}
\centering
  \includegraphics[width=\linewidth]{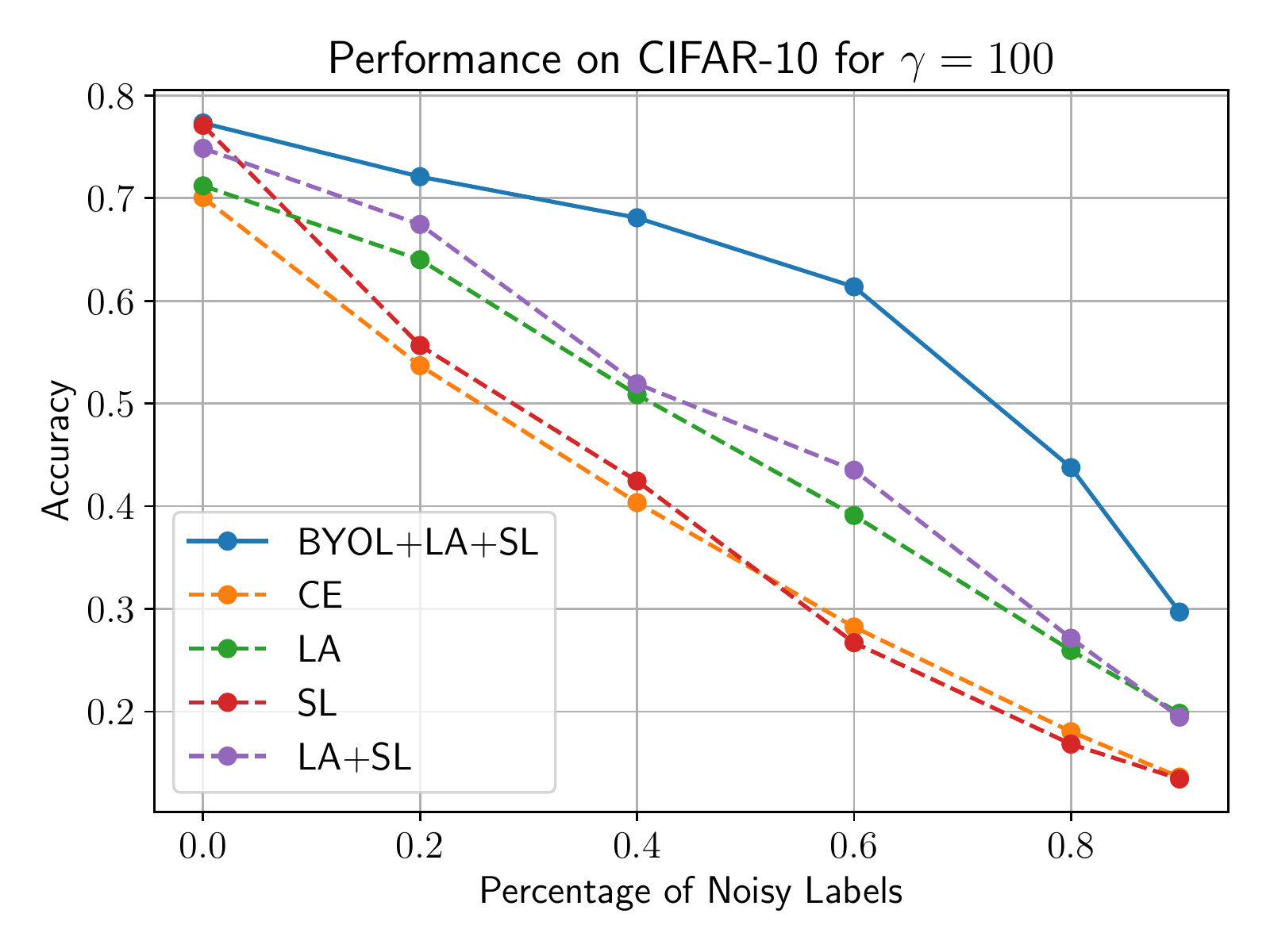}
\end{subfigure}
\centering
\begin{subfigure}{0.245\textwidth}
\centering
  \includegraphics[width=\linewidth]{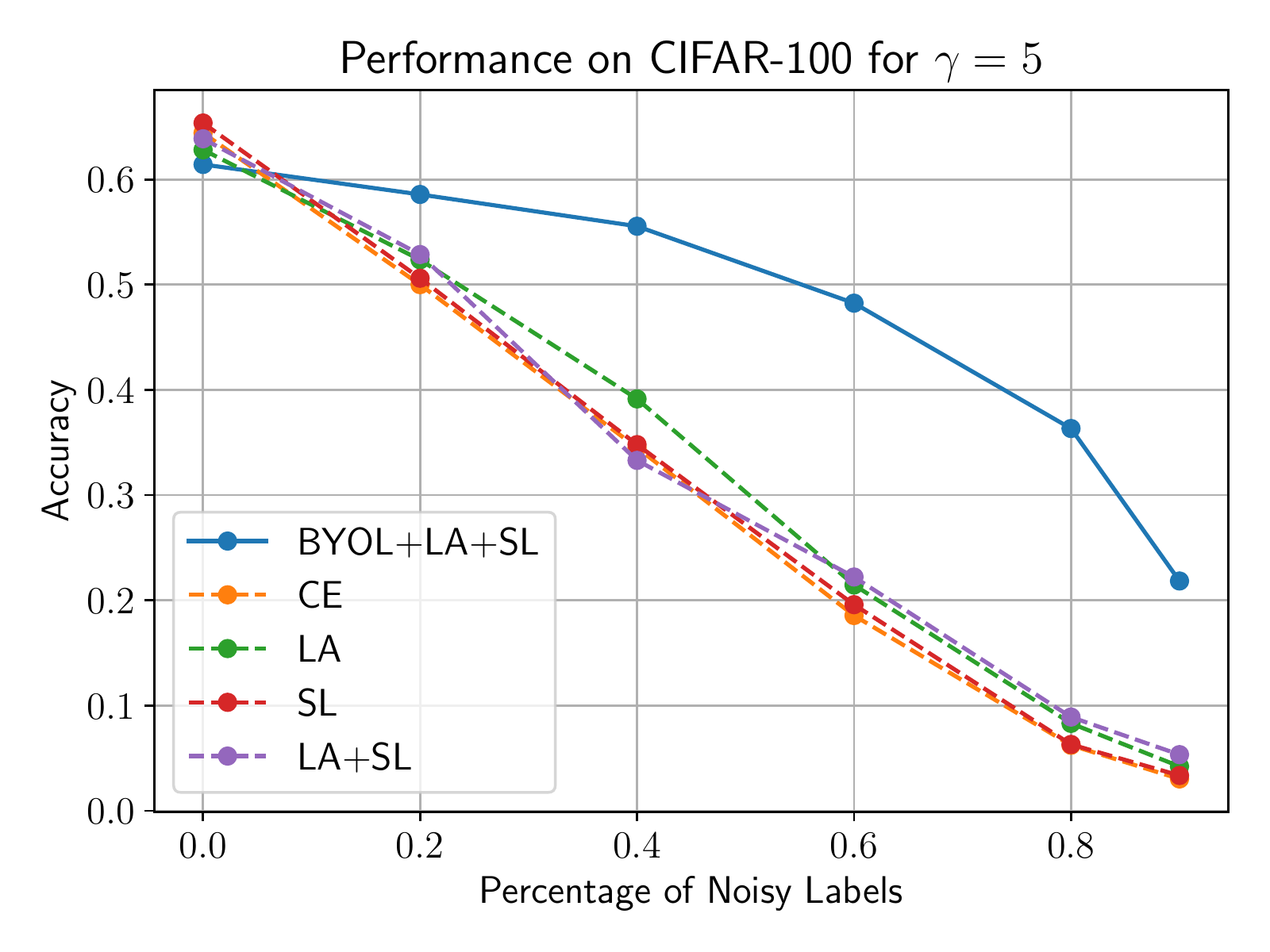}
\end{subfigure}
\begin{subfigure}{0.245\textwidth}
\centering
  \includegraphics[width=\linewidth]{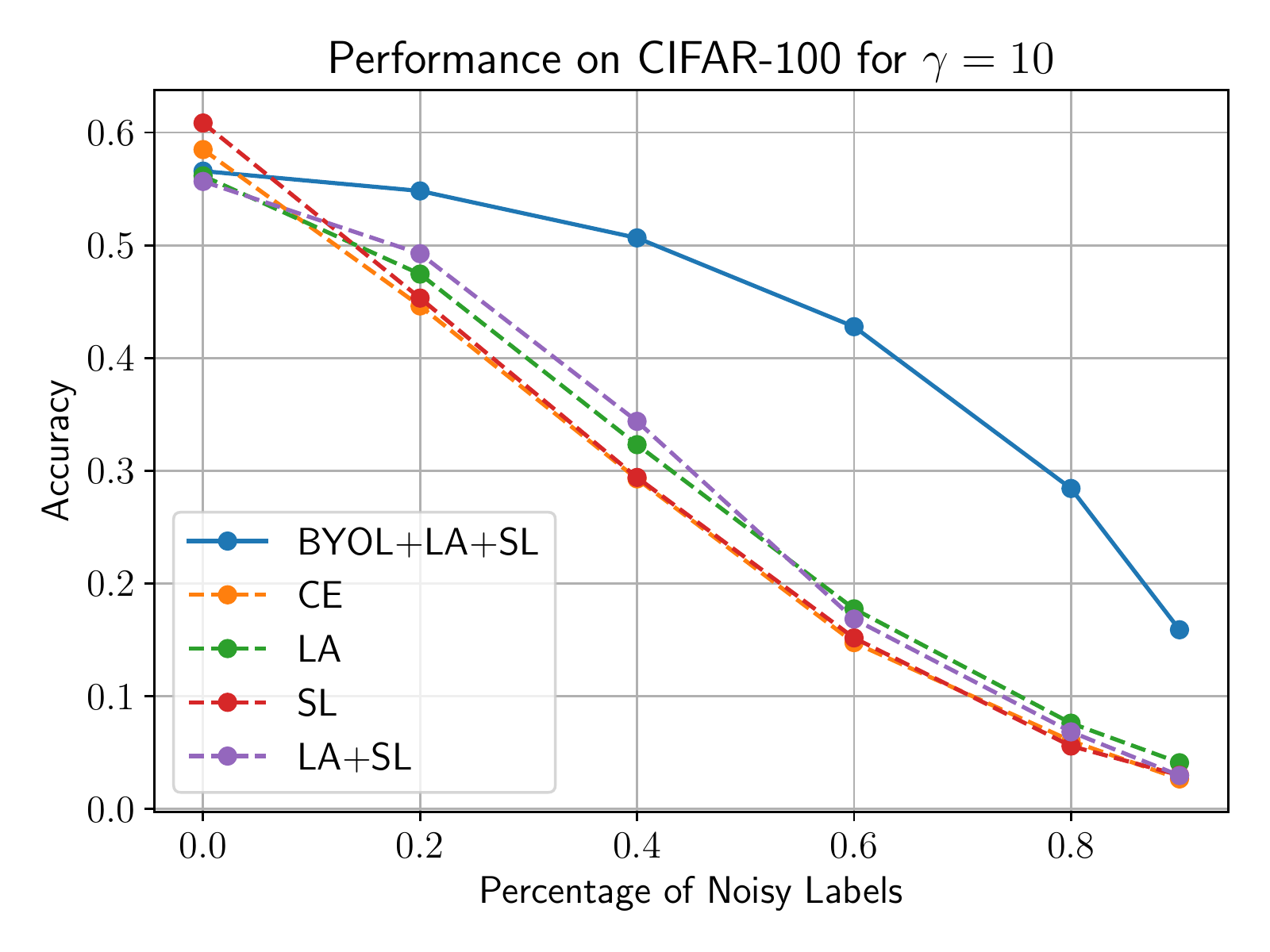}
\end{subfigure}
\caption{Comparison between two-stage training (here, BYOL with Logit-Adjusted and SuperLoss) 
    versus single-stage methods trained with various class imbalance- and noise-resistant losses
    (resp. Logit-adjusted and SuperLoss) on imbalanced versions of CIFAR-10 and CIFAR-100}
\label{fig:one_stage}
\end{figure*}

\textbf{Comparison with single-stage training.}
To measure how much two-stage self-supervised training is beneficial, we compare with networks trained from scratch in a single stage using the same losses.
This can be thought of an ablation study where we remove the self-supervised pre-training while keeping other things equal (\eg losses, number of epochs, etc). 
We compare single-stage learning results in Figure~\ref{fig:one_stage} with two-stage training using BYOL. 
In the absence of noise, we observe that single-stage training is on par or slightly better than two-stage training.
However, as soon as labels get noisy, two-stage training outperforms single-stage training by a large margin (\eg $+24\%$ for CIFAR-100 at 60\% noise and $\gamma=10$), even at very low noise levels. 
Interestingly, two-stage training requires only little more effort that single-stage training, as the second stage (\ie fine-tuning) is very short (10 epochs) and only updates a few layers.
This advocates for the use of SSL pre-training in imbalanced and possibly noisy situations, as it can provide state-of-the-art performance in difficult conditions.



\textbf{Comparison with SOTA.}
Experimental comparisons with state-of-the-art methods and other baselines are presented in Figure~\ref{fig:cifar_plots} for CIFAR-10 and CIFAR-100 at several imbalance levels.
Self-supervised pre-training leads to significant gains over state-of-the-art methods at any noise level in strongly imbalanced situations (\eg accuracy is 10\% to 30\% above for CIFAR-10 with $\gamma=100$). 

When the imbalance is moderate, self-supervised models are able to achieve reasonable performance (\eg 90.1\% to 92.0\% for CIFAR-10 when $\gamma=1$ and noise $\nu=20\%$), but they do not match the fully-supervised counterparts of DivideMix~\cite{li2020dividemix} and ELR~\cite{liu2020early}, which are able to achieve 95.7\% and 95.8\% in this setting, respectively.
Overall, these specialized approaches still perform better as long as both the imbalance and noise level are not too severe. 
For instance, all self-supervised methods start outperforming DivideMix on CIFAR-100 at $\gamma=5$ when the noise level is above~$70\%$.
More generally, we observe that the performance of the self-supervised models \emph{degrade much less}, even when the noise is increased to 80\% or 90\%. In particular, we achieve a new state-of-the-art for both CIFAR-10 and CIFAR-100 at 90\% noise by achieving (with SimSiam) 88.1\% and 53.0\% respectively in the imbalanced case. 
This is an improvement of 12.1\% and 21.5\% over DivideMix. 

\textbf{Discussion}. 
The poor performance achieved by ELR and DivideMix in the presence of class imbalance and noise is due to assumptions implicitly made in their design. Specifically, the regularization applied by ELR assumes that the clean samples are learnt first, followed by the noisy samples, and the regularization prevents the noisy samples from being memorized. 
However, in the imbalanced setting, the dominant classes are learnt first, followed by the classes on the long-tail. 
The regularization applied by ELR prevents the model from being able to learn the rare classes. 
In the case of DivideMix, there is a regularization term which encourages the model to predict a uniform class distribution, which is again violated in the long-tailed setting. 
In contrast, SimCLR, Barlow Twins, BYOL and particularly SimSiam are fairly robust to both the long-tailed distribution as well as the increasing label noise. 


\subsection{Clothing1M experiments}
\begin{figure*}[t]
\centering
\begin{subfigure}{0.4\textwidth}
\centering
\includegraphics[width=\linewidth]{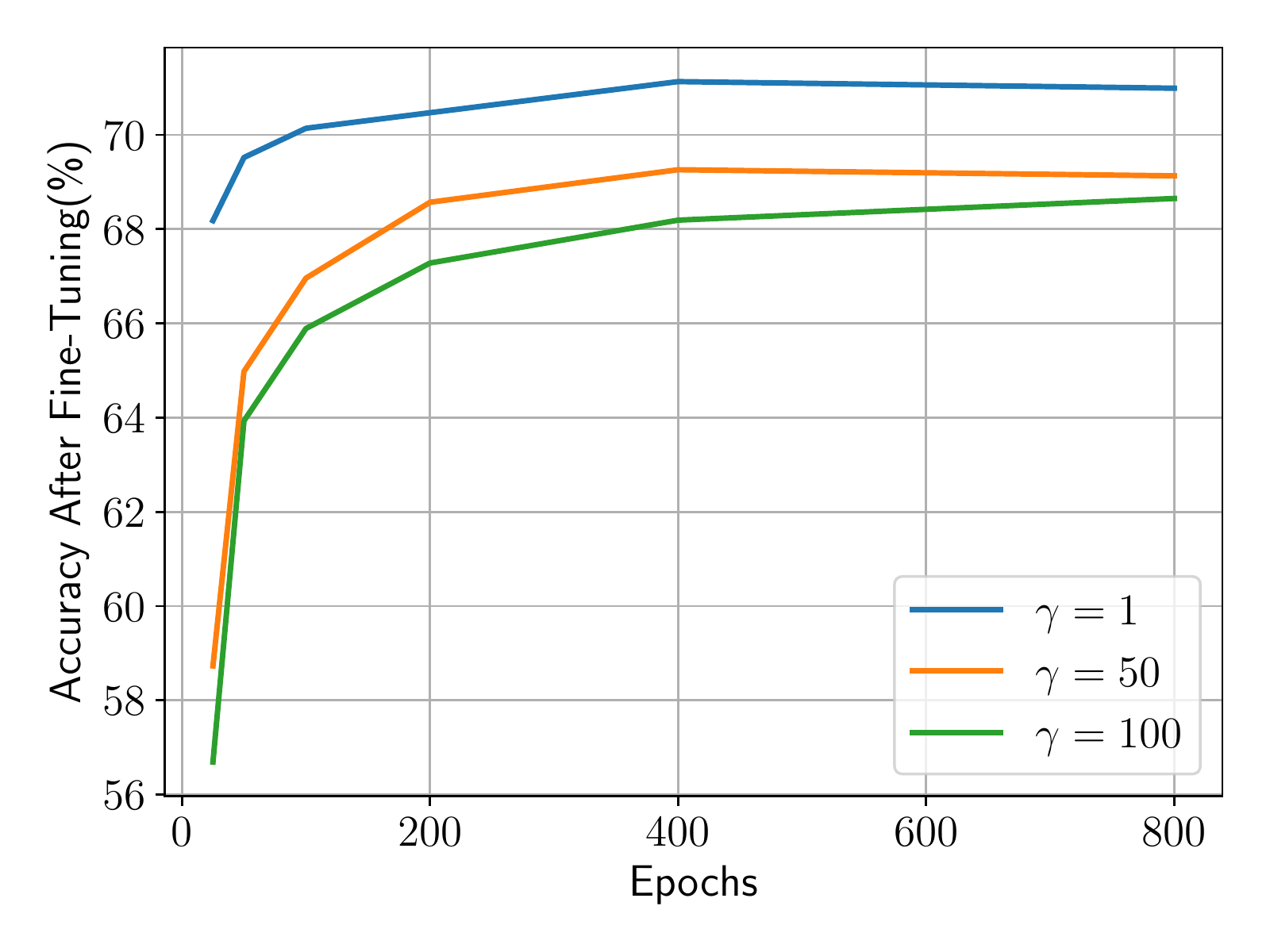}
\caption{}
\label{fig:c1epochs}
\end{subfigure}
~~~~~~~
\begin{subfigure}{0.4\textwidth}
\centering
  \includegraphics[width=\linewidth]{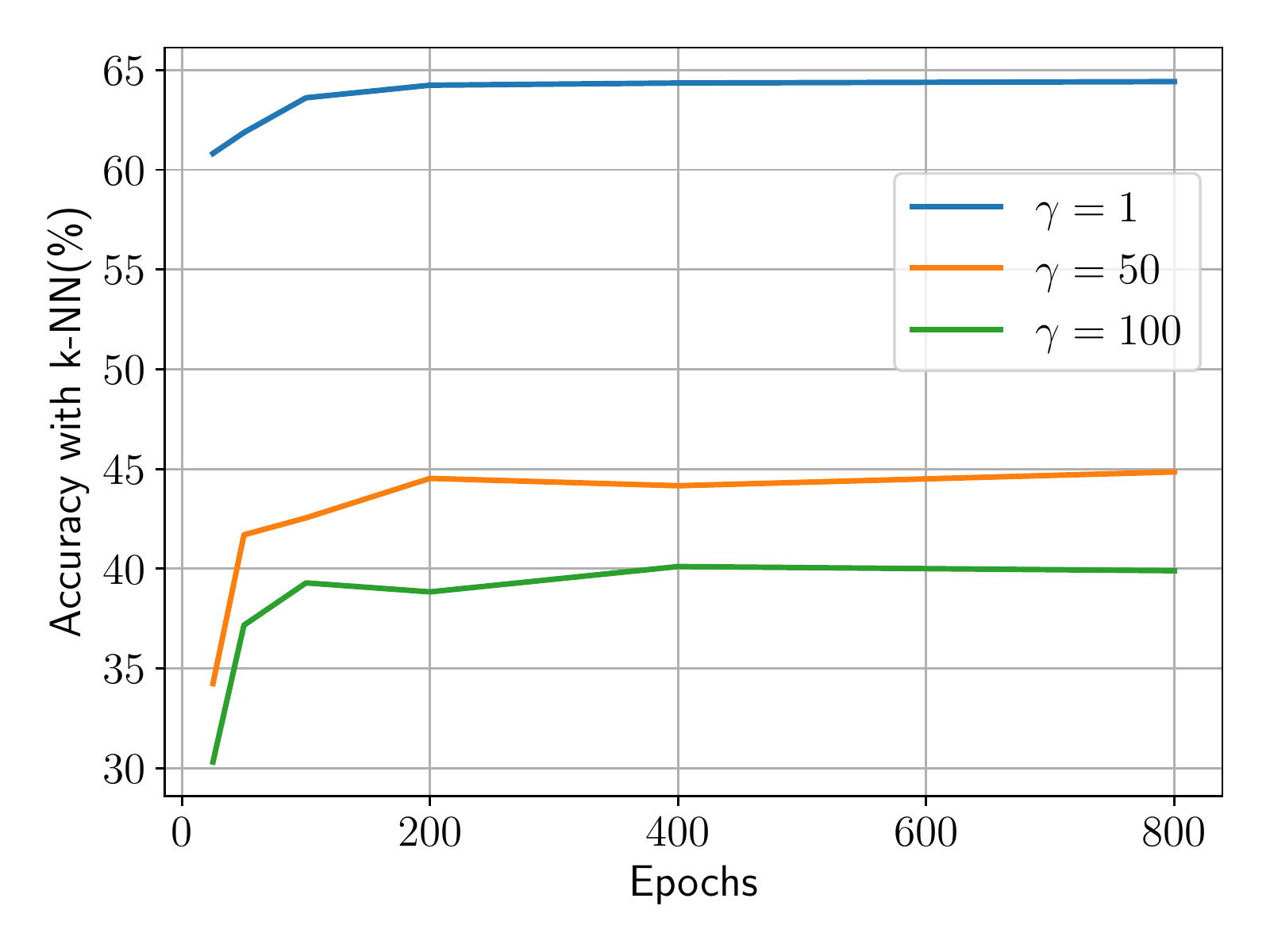}
  \caption{}
\label{fig:c1knn}
\end{subfigure}
\caption{Results for two-stage learning on the Clothing1M dataset as a function of the imbalance level and length of self-supervised pre-training (in epochs). For all $\gamma$ values, both the k-NN as well as the accuracy after fine-tuning values achieve their maximum values after 400 epochs.}
\label{fig:epochs}
\end{figure*}

We first evaluate the effect of the pre-training duration on the Clothing1M dataset in Figure~\ref{fig:c1epochs}. 
We also show the kNN-based proxy metric proposed by Chen~\etal~\cite{chen20simsiam} in Figure~\ref{fig:c1knn}. 
In imbalanced settings, we observe that it only weakly correlates with the much higher performance obtained after finetuning the model. 
In contrast to the kNN-based accuracy that rapidly stagnates, the actual accuracy after finetuning keeps increasing after 200 epochs, even though it gradually diminishes as is expected with this type of approach~\cite{chen20simsiam}.

\textbf{Comparison with SOTA.}
We benchmark our SimSiam-based self-supervised method on Clothing-1M to compare performance on a real-world noise model. 
Here, it is important to note that DivideMix and ELR use ImageNet initialization and model ensembling, which significantly contribute to their excellent performance. 
In comparison, our SimSiam-based model, trained from scratch and without any tricks, reaches an accuracy only 3\% below these more complex approaches.
Confirming earlier findings, we observe that their performance degrades sharply when imbalance is introduced. 
In contrast, our SimSiam model yields very similar performance regardless of the imbalance level, in accordance with earlier findings.
It significantly outperforms both DivideMix and ELR by more than 2\% and 4\% for $\gamma=5$ and $\gamma=10$, respectively. 
This shows that self-supervised pre-training is an effective strategy in large-scale datasets with realistic noise patterns.


\begin{table}[]
\vspace{-3mm}
\centering
\begin{tabular}{@{}lccc@{}}
\toprule
Method          & $\gamma=1$     & $\gamma=50$    & $\gamma=100$   \\ \midrule
DivideMix~\cite{li2020dividemix} & 73.9 & 67.1 & 64.9 \\
ELR~\cite{liu2020early}          & \textbf{74.2} & 63.9 & 59.6 \\
SimSiam+Logit+SuperLoss          & 71.1  & \textbf{69.3}  & \textbf{68.2} \\ \bottomrule
\end{tabular}
\vspace{1em}
\caption{Results on Clothing-1M with varying imbalance.}
\label{tab:c1m}
\vspace{-4mm}
\end{table}

\subsection{T-SNE projections}
\label{ssec:tsne}
\begin{figure}[]
\centering
\begin{subfigure}{0.3\textwidth}
\centering
\includegraphics[width=\linewidth]{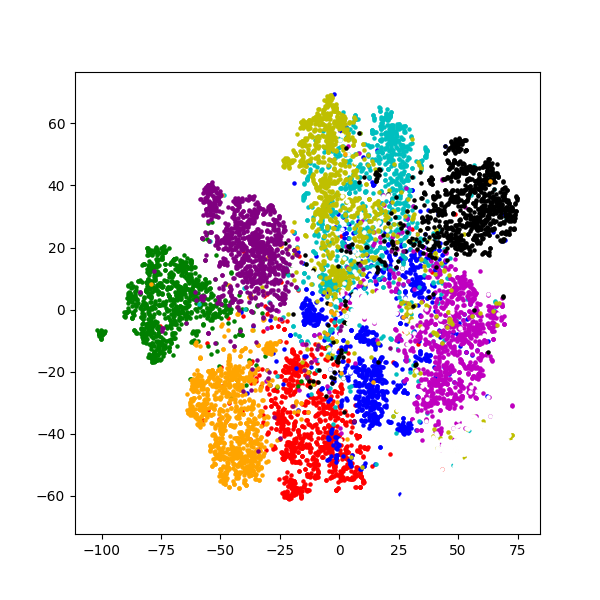}
\caption{}
\label{fig:6a}
\end{subfigure}
\begin{subfigure}{0.3\textwidth}
\centering
\includegraphics[width=\linewidth]{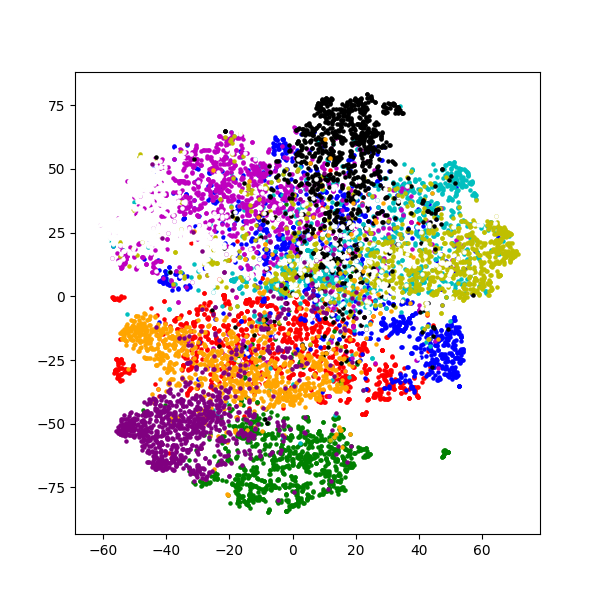}
\caption{}
\label{fig:6b}
\end{subfigure}
\begin{subfigure}{0.3\textwidth}
\centering
\includegraphics[width=\linewidth]{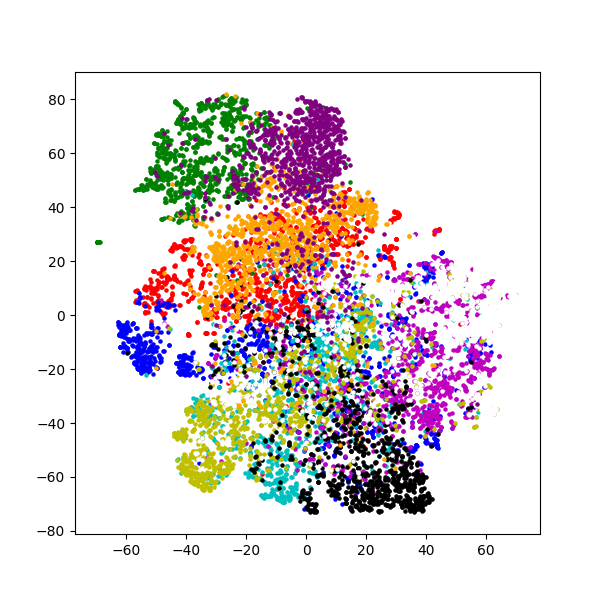}
\caption{}
\label{fig:6c}
\end{subfigure}
 \caption{t-SNE projections of SimSiam model on CIFAR-10, a) $\gamma=1$, b) $\gamma=50$, c) $\gamma=100$.}
\label{fig:tsne-cifar10} 
\end{figure}

We now illustrate the high quality of the representations learned via self-supervision in spite of class-imbalanced distributions.
In Figure~\ref{fig:tsne-cifar10}, we show t-SNE projections of representations learned using SimSiam on CIFAR-10 for several levels of imbalance; each projection plots all images from the balanced test set. 
As expected, we observe a good class separability in the balanced case. Nevertheless, severe class imbalance shows little impact to the separability of classes. They tend to stay well clustered, being it the dominant class in \textcolor{purple}{purple} or the 100 times smaller one in~\textcolor{red}{red}.
Overall, Figures~\ref{fig:6b} and \ref{fig:6c} show that the erosion of small class boundaries is rather limited compared to the balanced case.

\begin{figure*}[]
\centering
\begin{subfigure}{0.3\textwidth}
\centering
\includegraphics[width=\linewidth]{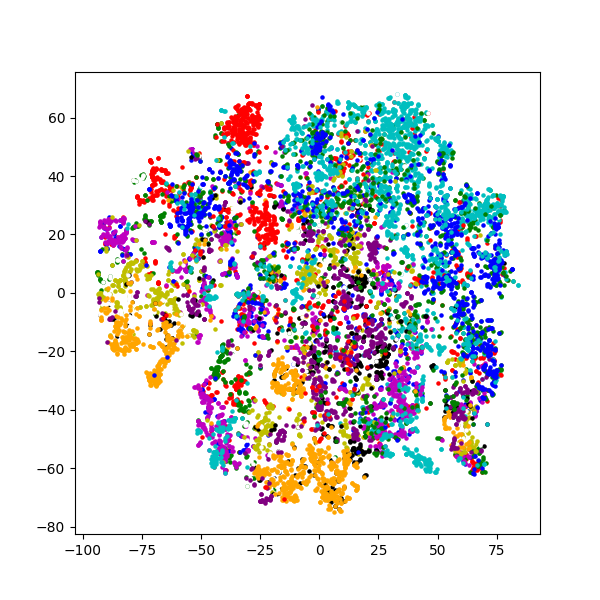}
\caption{}
\end{subfigure}
\begin{subfigure}{0.3\textwidth}
\centering
\includegraphics[width=\linewidth]{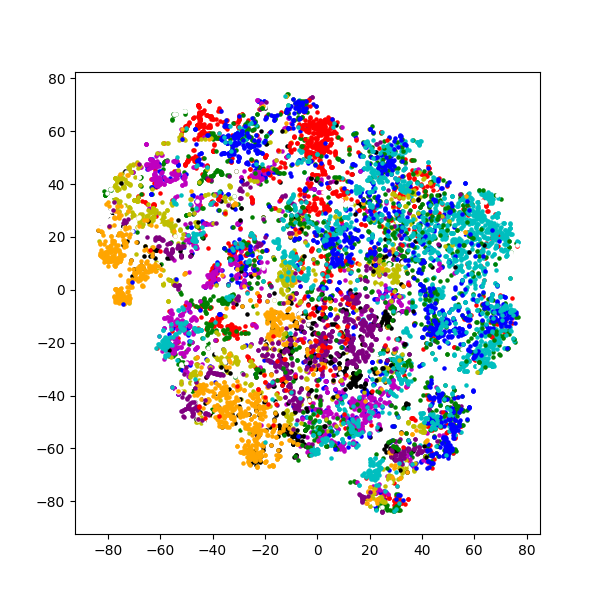}
\caption{}
\end{subfigure}
\begin{subfigure}{0.3\textwidth}
\centering
\includegraphics[width=\linewidth]{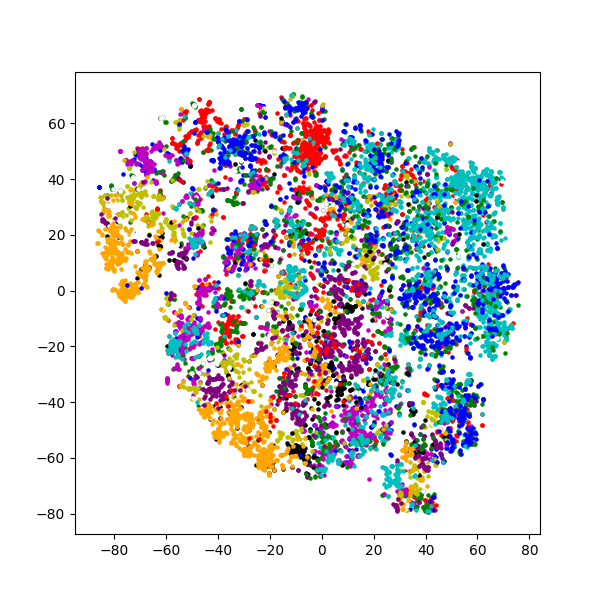}
\caption{}
\end{subfigure}
\caption{t-SNE projections of SimSiam model on Clothing-1M, a) $\gamma$=1, b) $\gamma$=50, c) $\gamma$=100.}
\label{fig:tsne-clothing1m}
\end{figure*}

Likewise, Figure~\ref{fig:tsne-clothing1m} shows t-SNE projections representations learned by SimSiam on Clothing-1M dataset for several imbalance levels. Note that classes in Clothing-1M are harder to separate, compared to the CIFAR-10 dataset. However, we again observe that severe class imbalance has little impact on the separability of classes. 
Here, the erosion of class boundaries for small classes is rather limited as the imbalance level increases.

\section{Related work}
\label{sec:sota}
We are not aware of any work that jointly tackles long-tailed learning and training with noise and review these two fields separately.

\textbf{Long-tailed learning}, 
\label{ssec:_sota_imbalance}
\ie learning with imbalanced classes, is a way to alleviate the performance degradation due to imbalance in the class distribution. 
Existing works in this field can be divided into three categories.
First, several methods focus on changing the \emph{data} that is given as input to the model~\cite{chawla2002smote,kubat1997addressing}. 
The most common strategy here is to over-sample the rare classes, or equivalently to under-sample the dominant classes.
A second solution is to modify either the algorithm design or the loss function that is used to train the model. 
Loss functions 
designed for imbalanced data include the focal loss~\cite{lin2017focal}, the logit adjustment loss~\cite{menon2020logitadj} as well as the label-distribution aware margin loss~\cite{cao2019learning}. Their common idea is to assign a higher loss to samples from rare classes, thereby providing a stronger supervisory signal for the model to learn these classes.
Thirdly, some methods modify the outputs predicted by the model. This family of post-hoc correction can either modify the threshold~\cite{menon2020logitadj}, or change the weights of the final classifier~\cite{kang2020decoupling} usually using some normalization procedure.

Decoupling the learning procedure into representation learning and classification was proposed for long-tailed data in~\cite{kang2020decoupling}. 
The key finding was that training a model using instance-balanced sampling conventionally, followed by training a classifier robust to imbalance works extremely well for long-tailed data. 
Our approach is very similar in spirit to this idea. 
However, in the presence of label noise, standard supervised learning collapses due to noise memorization. 
Therefore, we use self-supervised learning to learn effective representations, and train a classifier using losses that are robust to both data imbalance and label noise.  




\textbf{Learning with label noise}
\label{ssec:sota_noise}
is an active research field where existing approaches can be categorized into three groups.
First, \emph{label correction} methods aim to relabel the corrupted labels. 
They try to formulate explicit or implicit noise models to characterize the distribution of noisy and true labels~\cite{lee18cleannet}.
However, to recover the ground-truth labels, these approaches usually require the support of a small set of clean samples.
Second, \emph{loss correction} techniques seek to modify the loss function to achieve robustness, by using pre-calculated Backward or Forward noise transition matrix~\cite{patrini2017making},
or combining cross entropy and reverse cross entropy~\cite{wang19symmetric}.
Lastly, a third group of methods adopts \emph{sample selection} to identify potentially clean samples from a noisy training dataset. 
MentorNet~\cite{jiang18mentornet} introduces a data-driven curriculum learning paradigm in which a pre-trained mentor network guides the training of a student network. Co-teaching~\cite{han18coteaching} trains two DNNs simultaneously, and let them teach each other with some selected samples during every mini-batch. 
DivideMix~\cite{li2020dividemix}
trains two networks simultaneously and fits a Gaussian Mixture Model (GMM) on its per-sample loss distribution to divide the training samples into a labeled set and an unlabeled set. 

Early Learning Regularization (ELR)~\cite{liu2020early} is a recent advance in learning with noisy labels. 
The key observation here is that the clean samples are learnt first, followed by the noisy samples in the later epochs. 
Using this insight, ELR proposed a regularization term to prevent memorization of the noisy samples. 
All these methods are designed with the underlying assumption that classes are balanced.
As a result, this idea of separating noisy and clean samples using different techniques does not work as effectively when the data is imbalanced, as illustrated in our results in Section~\ref{sec:evaluation}.


\section{Conclusion}
\label{sec:conclusion}
In this work, we jointly tackle the problems of learning from long-tailed distributions and learning with noisy labels. 
Despite the vast literature that exists on both fields, these issues are usually tackled separately, often by making strong assumptions which are violated in the joint setting. 
Our proposed solution is inspired by recent findings in the field of semi-supervised learning.
It consists of a two-stage learning process that first pre-trains the model using one of the existing self-supervised techniques, followed by fine-tuning using a robust loss function.
We surprisingly find that all self-supervised methods that we experiment with are remarkably robust to class imbalance, even though they have not been explicitly designed for this use-case.
Overall, the proposed approach shows excellent robustness to both class imbalance and label noise, and set a new state of the art on CIFAR and on the real-world, large-scale dataset, Clothing1M in severe noise and class imbalance conditions. 
We hope that this serves as strong baseline for future exploration in this topic. 

\bibliographystyle{abbrv}
\bibliography{abbreviations,refs}


\end{document}